\pdfoutput=1
\documentclass[11pt]{article}

\usepackage[]{acl}

\usepackage{times}
\usepackage{latexsym}
\usepackage{xcolor}
\usepackage{lipsum}
\usepackage{microtype}

\usepackage{graphicx}
\usepackage{adjustbox}
\usepackage{amsmath}
\usepackage{amsfonts}
\usepackage{bm}
\usepackage{dsfont}
\usepackage{soul}
\usepackage[ruled, noend]{algorithm2e}
\usepackage{amsthm}
\usepackage{enumitem}
\usepackage{booktabs}
\usepackage{multirow}
\usepackage{listings}
\usepackage{subcaption}
\usepackage{todonotes}
\usepackage{siunitx}

\makeatletter
\newcommand\notsotiny{\@setfontsize\notsotiny{6.31415}{7.1828}}
\makeatother

\newcommand{\docm}{$\theta^{\mathrm{doc}}$}
\newcommand{\senm}{$\theta^{\mathrm{sen}}$}
\newcommand{\augm}{$\theta^{\mathrm{augdoc}}$}
\newcommand{\zaugm}{$\theta^{\mathrm{augdoc}}_{\mathrm{zero-shot}}$}
\newcommand{\dtrainsen}{$\mathcal{D}_{\mathrm{train}}^{\mathsf{sen}}$}
\newcommand{\dtraindoc}{$\mathcal{D}_{\mathrm{train}}^{\mathsf{doc}}$}
\newcommand{\dvalsen}{$\mathcal{D}_{\mathrm{val}}^{\mathsf{sen}}$}
\newcommand{\dvaldoc}{$\mathcal{D}_{\mathrm{val}}^{\mathsf{doc}}$}
\newcommand{\dtestsen}{$\mathcal{D}_{\mathrm{test}}^{\mathsf{sen}}$}
\newcommand{\dtestdoc}{$\mathcal{D}_{\mathrm{test}}^{\mathsf{doc}}$}

\def\parsum#1{\bgroup \textcolor{blue}{Paragraph summary: #1}\egroup}

\def\bluenote#1{\bgroup \textcolor{blue}{#1} \egroup}
\def\cyannote#1{\bgroup \textcolor{cyan}{#1} \egroup}
\def\tealnote#1{\bgroup \textcolor{teal}{#1} \egroup}
\def\ronenote#1{\bgroup \textcolor{olive}{R1: #1} \egroup}
\def\rtwonote#1{\bgroup \textcolor{red}{R2: #1} \egroup}
\def\rthreenote#1{\bgroup \textcolor{orange}{R3: #1} \egroup}

\title{Granularity is crucial when applying differential privacy to text: An investigation for neural machine translation}

\author{Doan Nam Long Vu\textsuperscript{$1$}
	\and Timour Igamberdiev\textsuperscript{$1$}
	\and \textbf{Ivan Habernal}\textsuperscript{$2$} \\
	Trustworthy Human Language Technologies \\
	\textsuperscript{$1$} Department of Computer Science, Technical University of Darmstadt \\
	\textsuperscript{$2$} Research Center Trustworthy Data Science and Security of the University Alliance Ruhr,\\
	Faculty of Computer Science, Ruhr University Bochum \\
	{\texttt{timour.igamberdiev@tu-darmstadt.de}} \\
	\url{www.trusthlt.org}
}

\begin{document}
    \onecolumn
    \noindent \textbf{Granularity is crucial when applying differential privacy to text: An investigation for neural machine translation}

    \medskip
    \noindent Doan Nam Long Vu, Timour Igamberdiev, Ivan Habernal

    \bigskip
    This is a \textbf{camera-ready version} of the article accepted for publication at the \emph{Findings of the Association for Computational Linguistics: EMNLP 2024}. The final official version will be published on the ACL Anthology website later in 2024: \url{https://aclanthology.org/}

    \medskip
    Please cite this pre-print version as follows.
    \medskip

\begin{verbatim}
@InProceedings{Vu.et.al.2024.EMNLP,
    title = {Granularity is crucial when applying
             differential privacy to text: An investigation
             for neural machine translation},
    author = {Vu, Doan Nam Long and Igamberdiev, Timour and
              Habernal, Ivan},
    publisher = {Association for Computational Linguistics},
    booktitle = {Findings of the Association for 
                 Computational Linguistics: EMNLP 2024},
    pages = {(to appear)},
    year = {2024},
    address = {Miami, United States}
}
\end{verbatim}
    \twocolumn

\maketitle
\begin{abstract}
Applying differential privacy (DP) by means of the DP-SGD algorithm to protect individual data points during training is becoming increasingly popular in NLP.
However, the choice of granularity at which DP is applied is often neglected.
For example, neural machine translation (NMT) typically operates on the sentence-level granularity. From the perspective of DP, this setup assumes that each sentence belongs to a single person and any two sentences in the training dataset are independent. This assumption is however violated in many real-world NMT datasets, e.g., those including dialogues. For proper application of DP we thus must shift from sentences to entire documents.
In this paper, we investigate NMT at both the sentence and document levels, analyzing the privacy/utility trade-off for both scenarios, and evaluating the risks of not using the appropriate privacy granularity in terms of leaking personally identifiable information (PII).
Our findings indicate that the document-level NMT system is more resistant to membership inference attacks, emphasizing the significance of using the appropriate granularity when working with DP.\footnote{Our code is available at \url{https://github.com/trusthlt/granularity-is-crucial-dp}.}
\end{abstract}

\section{Introduction}
\label{sec:introduction}

With increasing concerns about the privacy of individuals and data leakage from NLP systems~\citep{carlini2021extracting},
a method that has gained popularity in privacy-preserving NLP is Differential Privacy (DP)~\citep{hu-etal-2024-differentially}. However,
the exact manner in which DP is applied to a textual dataset has numerous pitfalls. The \textit{unit of privacy} is one among them, i.e.\ the granularity at which we assume an individual `data point' (e.g.\ sentences, documents, and so forth)~\citep{ponomareva-etal-2022-training,igamberdiev-habernal-2023-dp}, 
with an assumption of \textit{independence} among data points~\citep{Dwork.Roth.2013}.

One particular task that has recently raised many privacy concerns is neural machine translation (NMT).
Applying DP at the sentence level for NMT may break the independence assumption if more than one sentence is associated with a single individual~\citep{brown2022does}, as depicted in \autoref{fig:example-iid}.
In such cases, scaling up the unit of privacy to the document level by grouping related sentences overcomes the violated privacy protection which `pretends' all sentences are independent, i.e. the status quo.

\begin{figure}[!tb]
	\centering
        \begin{tabular}{l}
        \begin{lstlisting}[escapechar=$, label={lst:iid-maia}, basicstyle=\notsotiny, upquote=true]
{
...
"de": "Kunde:$\colorbox{green}{Immo Hande-Hornig}$",
"en": "Customer: Immo Hande-Hornig",
...
"de": Agent: ... Ich bin$\colorbox{green}{Immo Hande-Hornig}$.
"en": Agent: ... you are through to Immo Hande-Hornig.
...
}
            \end{lstlisting}
         \end{tabular}
     \caption{Examples of sentences that are \textit{not independent} within a document. The independence~is violated via the ``\colorbox{green}{Immo Hande-Hornig}'' sequence, breaking the DP guarantee of protecting each sentence during the training process.}
    \label{fig:example-iid}
\end{figure}

The main objective of this paper is to compare the use of DP for NMT systems and datasets at the sentence and document levels, focusing on the level of privacy protection offered.
First, we investigate the trade-off between privacy and utility for different levels of granularity (sentence vs.\ document) during the training process of an NMT system.
Specifically, we examine how performance is affected when applying DP with varying levels of privacy guarantees and granularity. 
Secondly, we aim to evaluate the risks of not using a proper privacy granularity during the training process of an NMT system through data extraction attacks on these systems.

Our contributions are as follows.
(1) We propose a novel approach to apply differential privacy to NMT systems at the document level, utilizing the DP-NMT framework~\citep{igamberdiev-habernal-2023-dp} and the mLongT5 model~\citep{uthus_mlongt5_2023}.
The evaluation results show that the document-level NMT system is extremely sensitive to the privacy budget ($\varepsilon$), which can significantly affect performance and lead to a drop in utility.
We therefore suggest training the document-level NMT system on a larger, non-sensitive dataset, such as WMT22~\citep{kocmi-etal-2022-findings}, to achieve a better trade-off between privacy and utility on the downstream dataset with DP\@.
(2) We apply the loss-based membership inference attack (MIA)~\citep{yeom-2018} to detect private information in NMT systems at both the sentence and document levels.
Based on this MIA, we create an evaluation schema for personally identifiable information (PII) to estimate the percentage of potential information leakage.
Our results show that the document-level NMT system is more robust against the loss-based MIA than the sentence-level NMT system, demonstrating the importance of using the proper granularity when working with DP.

\section{Background}
\label{sec:background}

\subsection{Differential Privacy}

We refer to
Appendix~\ref{sec:dp} for 
a detailed explanation of
Differential Privacy and DP-SGD.

A key aspect of applying DP to text 
is that we cannot simply utilize group privacy~\citep{Dwork.Roth.2013} to achieve document-level privacy from sentence-level privacy.
Since DP-SGD leverages the 
Approximate DP definition~\citep{Dwork.Roth.2013}, the $\delta$ value, 
i.e. the probability of privacy leakage occurring, 
is not $0$ when applying group privacy.
If the number of data points $k$ in which two neighboring datasets differ is large enough, $\delta$ will exceed $1$ due to scaling with a factor of $k e^{(k-1)\varepsilon}$.
Moreover, using relaxed DP definition such as Renyi DP~\citep{mironov_renyi_2017} to avoid including the $\delta$ value will scale $\varepsilon$ to a very large value that is practically unmanageable, also resulting in a very weak privacy guarantee.

\subsection{Overview of membership inference attacks (MIA)}
Since datasets used to train neural models
often contain confidential user information, they may be vulnerable to privacy risks~\citep{hu_membership_2022}.
The trained models are often over-paramet erized, meaning they can memorize information about their training dataset~\citep{mireshghallah-etal-2022-empirical}.
They exhibit a different behavior on training data compared to test data, 
with model parameters storing information about specific training data unit.
Membership inference attacks (MIAs) aim to predict whether specific examples are members of the training dataset~\citep{hu_membership_2022}.
In general, an MIA is actually a binary classifier, which is designed to distinguish a target model's behavior of its training members from the non-members.

The first MIA was proposed by~\citet{shokri_membership_2017}, 
utilizing shadow datasets that have similar distribution to the original training data.
Multiple shadow models are trained on these datasets, which are meant to mimic the behavior of the target model.
The output predictions of these models are then used as input to a final binary classification model.
This model detects whether a given data point belongs to the target model or not.

\section{Related Work}
\label{sec:related-work}
\subsection{Previous work on DP+NLP}
Several works attempted to pre-train language models with DP-SGD
~\citep{anil-etal-2022-large, yin-habernal-2022-privacy, ponomareva-etal-2022-training}.
With a significant computational burden of pre-training with DP-SGD, finetuning
pre-trained language models using DP-SGD has seen increased research over the past few years~\citep{Senge.et.al.2022.EMNLP, li2022large}.
The main objective is to utilize a pre-trained checkpoint of a model that was created by using a publicly available corpus of data, such as Wikipedia or C4~\citep{colin-2020-t5} and then fine-tune it on a private downstream dataset with DP-SGD.
Although fine-tuning is more efficient than pre-training, it still requires a large amount of data on text generation tasks to achieve good performance, e.g.\ language modeling~\cite{li2022large} or NMT~\citep{igamberdiev-etal-2024-dp}.
The aforementioned related works only concentrate on the protection of the gradient at the \textit{sentence level}. In contrast, the current work is concerned with the protection of the gradient at the \textit{document level}.

\subsection{Previous work on document-level NMT}
Sentence-level NMT is the most common method of machine translation because it is simpler to train and evaluate with existing large datasets and evaluation metrics~\citep{post_escaping_2023}.
The primary limitation is due to the NMT model's memory consumption for sequence length, resulting in a larger memory footprint when increased. 
Secondly, popular datasets, such as WMT~\citep{fernandes_measuring_2021}, exist only for sentence-level machine translation, even though they were originally created as documents.
Nonetheless, this approach has its limitations, as demonstrated in the current study,
which is the related privacy issue with DP-SGD at the sentence level.
Moreover, from the perspective of machine translation, a sentence-level MT model is not suitable for translating lengthy documents without taking their context into account~\citep{wicks_identifying_2023, wu_document_2023}.
Recent related works on document-level machine translation can be separated into two categories: \textit{Encoder-Decoder Models} and \textit{Decoder Only Models}.
\paragraph{Encoder-Decoder Models}
Most of the recent works focus on the standard Transformer model~\citep{wu_document_2023, zhuocheng-etal-2023-addressing}. Typically, they concatenate multiple sentences to form a document with the length up to 512 or maximum 1024 tokens for training. However, the naive approach generally suffers from a \textbf{length bias problem}, which causes significant degradation in translation quality when decoding documents that are much shorter or longer than the maximum sequence length during training~\cite{zhuocheng-etal-2023-addressing}.
\paragraph{Decoder Only Models}
Unlike the vast majority of the training/fine-tuning paradigm, recent works~\citep{hendy_how_2023, karpinska-iyyer-2023-large} suggest that Generative Pre-Training (GPT) models ~\citep{radford2018improving} are able to achieve very competitive translation quality on document-level translation.
As such, they use a few-shot prompting technique to translate a document, employing ChatGPT\footnote{\url{https://openai.com/blog/chatgpt}}.
The prompt displays examples of each translated sentence pair first and instructs the model to consider the context when translating, as in a document.

\section{Methods}
\label{sec:methods}

\subsection{Document-level machine translation}
We employ the DP-NMT framework developed by ~\citet{igamberdiev-etal-2024-dp} for privacy-preserving NMT with DP-SGD\@.
The framework is built on top of Flax~\citep{flax2020github} and JAX~\citep{jax2018github} for rapid DP-SGD training.
By default, the framework supports models such as mBART~\citep{liu2020multilingual} or mT5~\citep{xue-etal-2021-mt5} out of the box.
However, these multilingual seq2seq models are not suitable for our task, as they are pre-trained on shorter 
sentences.

\paragraph{mLongT5 model}

mLongT5 is a multilingual seq2seq model 
based on LongT5~\citep{guo_longt5_2022}, which is a seq2seq model that uses T5~\citep{colin-2020-t5} as its foundation, with a \textit{Transient Global ~(TGlobal) Attention} mechanism. This attention mechanism is well-suited for long text tasks in terms of memory efficiency, and mLongT5 leverages the mC4 dataset \citep{xue-etal-2021-mt5} 
with 4096 token long input sequences for pre-training.
As mLongT5 checkpoints are designed for long text tasks and available for Flax and JAX, we incorporate mLongT5 into the DP-NMT framework.

\subsection{Loss-based MIA}

Previous work attacking an NMT system used the entire WMT dataset to create a sophisticated shadow MIA attack~\citep{hisamoto2020membership}.
However, in our work, the datasets on which we conduct experiments (see Section~\ref{sec:experiments}) have less than 20,000 data points, which is significantly fewer than the millions of data points in WMT. This makes shadow MIA less effective. 
Also, in terms of computational complexity for the shadow MIA, an attacker must train hundreds of shadow models to achieve good performance~\citep{yeom-2018}.
Loss-based metrics~\cite{yeom-2018} are less computationally intensive to perform.
Intuitively, if the loss of a data point is smaller than the target model's expected training loss, the record is classified as a member; otherwise as a non-member.
The target model is trained by minimizing the prediction loss of its training members.
Therefore, the prediction loss of a training record should be smaller than that of a test record.
The attack $\mathsf{Exp}^{\mathsf{M}}_{\mathsf{loss}}$ is defined as follows:

\begin{equation}
    \mathsf{Exp}^{\mathsf{M}}_{\mathsf{loss}} = \mathds{1}(\ell( \theta(\mathbf{r} |\mathbf{s} ); \mathbf{r}) \leq \tau), \label{eq:equation}
\end{equation}
where $\theta$ is the model, $\mathbf{r}$ is target output, $\mathbf{s}$ is the input, $\ell$ is the loss function (typically cross-entropy loss), $\tau$ is a threshold (average training loss) and $\mathds{1}$ is a classifier function, which takes event $A$ and returns 1 if the event $A$ occurs, 0 otherwise.

Loss-based MIA highlights that \textbf{overfitting} of target ML models is the primary factor contributing to the success of MIAs.
This attack \textit{strongly exploits} the different behaviors of target ML models on their training versus test data.
The attacker is assumed to have knowledge of the data points, but it is uncertain whether they were used in training.
In fact, we are assuming a very powerful adversary that already has knowledge of the data, which is why this is a strong white-box attack for investigating data leakage.
The original evaluation metric scheme for the loss-based MIA\@ is the attack advantage or privacy leakage resulting from 
differences between the false positive rate (FPR) and true positive rate (TPR) of the attack:
    $\mathsf{Adv}^{\mathsf{M}} = \mathrm{TPR} - \mathrm{FPR}$

\subsection{PII exposure}
Considering \textit{private information} as named entities that a model might overfit to during the training process,
we present a \textit{evaluation scheme} to test the effectiveness of the MIA with respect to PII.
The method works as follows: We carry out the loss-based MIA, obtaining extracted training records from Eqn.~\ref{eq:equation}.
We then select the \textbf{true positive predictions} and calculate the number of PII that are present.
For sentence-level privacy, due to correlation of sentences within a document, more PII leakage would be expected than with document-level privacy.
This aims to see how well the model optimizes against entities.
We use the default pipeline setting of \texttt{Presidio}\footnote{\url{https://microsoft.github.io/presidio/}} which consists of RegEx, Spacy NER and BERT contextual awareness to extract a set of PII from given sentences.\footnote{We note that \texttt{Presidio} is not as thorough as human annotators, and may miss some PII data or return false positives. However, it is still a good approximation of the number of PII.}

\section{Experiments}\label{sec:experiments}
\subsection{Datasets}
We aim to find a suitable dataset which mimics the real private environment of processing sensitive information, but is publicly available, both for reproducibility and ethical reasons. In addition, a dataset must be appropriate 
for both sentence-level and document-level machine translation. Therefore, we select three datasets for our investigations, BSD, MAIA and Europarl, described below.

 \paragraph{BSD} The Business Scene Dialogue corpus (BSD)~\citep{rikters2019designing} is a collection of fictional business conversations in various scenarios,
 with parallel data for Japanese and English.
 For our experiments, we combined the original corpus, which consists of two translation direction
 into a single Japanese $\to$ English (JA-EN) language pair.

\paragraph{MAIA} The Multilingual Artificial Intelligence Agent Assistant (MAIA) corpus consists of genuine bilingual (German-English) customer support conversations 
from the Unbabel database~\citep{farinha-etal-2022-findings}.
To make the conversations publicly available, the data was first anonymized using the Unbabel proprietary anonymization tool and then manually verified.

\paragraph{Europarl}

The Europarl~\citep{koehn-2005-europarl} dataset is a widely used parallel corpus in the field of machine translation. Extracted from the proceedings of the European Parliament, the dataset includes versions in 21 European languages. For this work, we use the Europarl V10 version from WMT22~\cite{kocmi-etal-2022-findings} for German-English bilingual text.

\begin{table}[!t]
\resizebox{\columnwidth}{!}{
\begin{tabular}{@{}llrrrr@{}}
\toprule
\textbf{Dataset} &  \textbf{Level}    & \textbf{\# Train}  & \textbf{\# Val.} & \textbf{\# Test} \\ \midrule
\multirow{2}{*}{BSD}&  Sentence & 20,000 & 2,120 & 2,051 \\
                      & Document & 670   & 69   & 69   \\
\multirow{2}{*}{MAIA}& Sentence & 13,380 & 2,488 & 2,109 \\
                      & Document & 355   & 71   & 70\\
\multirow{2}{*}{Europarl}& Sentence & 1,454,229 & 181,774 & 181,764 \\
                      & Document & 143,706   & 19,786   & 19,967\\
    \bottomrule
\end{tabular}}
\caption{Number of training examples for both datasets in sentence-level and document-level}
\label{tab:table-data}
\end{table}

\subsection{Data preparation}
Since those datasets are dialogue/speech session datasets, we need to concatenate the utterances within a dialogue/speech session into a single document for document-level machine translation.
First, we concatenate the speaker's name and the utterance into a single sentence.
Namely, \texttt{<SPEAKER>: <UTTERANCE>}.
Then we concatenate all the utterances within a dialogue/speech session into a single document.
This process results in a smaller number of training examples than the original sentence-level training examples.
\autoref{tab:table-data} shows the number of training examples for each datasets at the sentence level and document level.
We refer to Figures~\ref{fig:example-doc-vs-sen-maia},~\ref{fig:example-doc-vs-sen-bsd} and \autoref{fig:example-doc-vs-sen-europarl}
in \autoref{sec:data_prep} for examples of each dataset preparation. 

\subsection{Experimental setup}

The training experiment directly on Huggingface's mLongT5 checkpoint\footnote{\url{https://huggingface.co/agemagician/mlong-t5-tglobal-base}} is denoted as~\senm~at the sentence level, and as~\docm~at the document level.
We denote training data at the sentence level as~\dtrainsen~and at the document level~as~\dtraindoc, similarly for validation~(\dvalsen,~\dvaldoc)~and test data~(\dtestsen,~\dtestdoc).
We refer to \autoref{tab:notation} in Appendix for the notation and its description used in this work.

\paragraph{Additional pre-training with WMT22}
The number of document-level training examples is much smaller than at the sentence level (See~\autoref{tab:table-data}).
Thus, the model may underfit the training data during private training, resulting in poorer performance than normal training at the document level and private training at the sentence level.
To improve the document-level model performance during private training, we fine-tuned mLongT5 checkpoint without DP-SGD on the WMT22 dataset first before fine-tuning on downstream datasets at the document level.
After fine-tuning on the document-level WMT22 dataset, we fine-tune the model on the BSD and MAIA datasets at the document level for both normal and private training. We also use the pre-trained checkpoint on WMT22 with Europarl for comparison.
The model training experiments based on the document-level WMT22 dataset are denoted as~\zaugm~and with downstream data~as~\augm.

\begin{figure*}[!ht]
\centering
\begin{subfigure}{.33\textwidth}
  \centering
  \includegraphics[width=\linewidth]{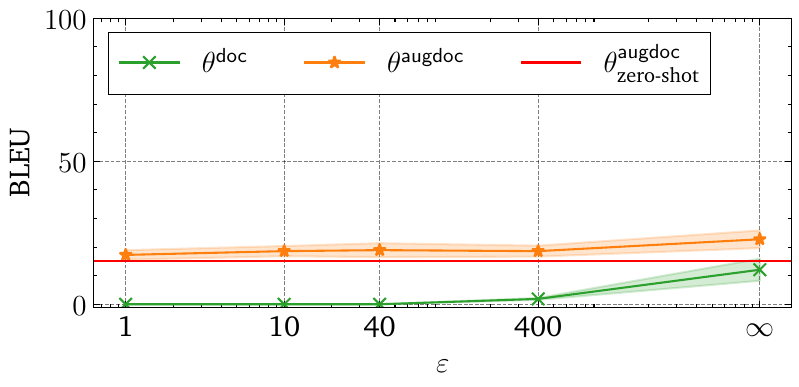}
  \caption{BSD}
  \label{fig:bleu-doc-test-bsd}
\end{subfigure}
\begin{subfigure}{.315\textwidth}
  \centering
  \includegraphics[width=\linewidth]{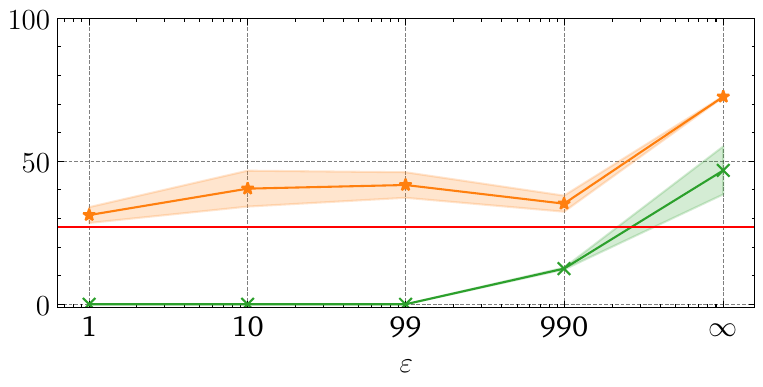}
  \caption{MAIA}
  \label{fig:bleu-doc-test-maia}
\end{subfigure}%
\begin{subfigure}{.315\textwidth}
  \centering
  \includegraphics[width=\linewidth]{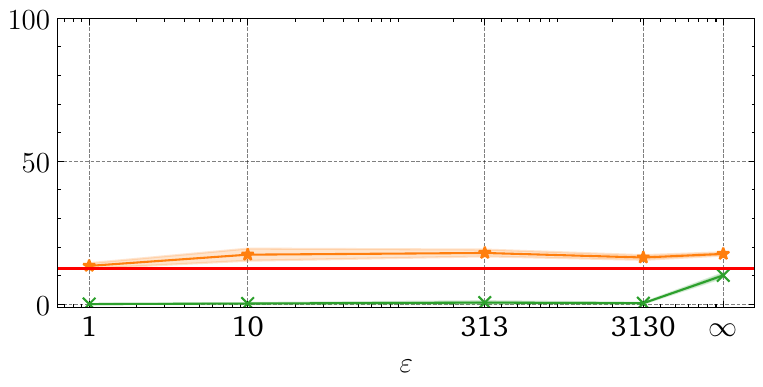}
  \caption{Europarl}
  \label{fig:bleu-doc-test-europarl}
\end{subfigure}%
\caption{BLEU scores on \dtestdoc~for the three document-level model fine-tuning configurations. Lower $\varepsilon$ corresponds to better privacy.}
\label{fig:bleu-doc-test}
\end{figure*}

\subsection{Hyperparameters}
The primary distinction between two level models in terms of hyperparameters is the maximum sequence length. For sentence-level training, it is set to 64-128, whereas for document-level training, it is set to 1200-1500. 
We refer to the details of our hyperparameters search in \autoref{sec:hyperparameters}. 

For additional training with WMT22, we use the same hyperparameters as document-level settings. To prepare the document-level WMT22 dataset, we concatenate the multiple sentences into a single document, as long as they reach 1200 tokens for Japanese to English and 1600 tokens for German to English.
Those documents are aligned with the original sentence-level training examples.
We refer to \autoref{tab:tablewmt} in Appendix~\ref{subsec:aug_training} for the number of training examples in our experiment with WMT22.

\paragraph{Privacy Hyperparameters}
We compare $\varepsilon$ values\footnote{The maximum number of utterances in a dialogue/speech session within each dataset (99 for MAIA and 40 for BSD, 313 for Europarl) is multiplied by the $\varepsilon$ values of 1 and 10. Finally, the resulting values are $\varepsilon$ equals 990, 90 for MAIA, then 400, 40 for BSD and 3130, 313 for Europarl.} of $\infty, 990, 90, 10$ and $1$ for training on MAIA, then $\infty, 400, 40, 10$ and $1$ for training on BSD and  $\infty, 3130, 313, 10$ and $1$ for training on Europarl.
Those values are applied to both sentence-level and document-level training.
We refer to the details of our privacy guarantee in \autoref{sec:our-privacy-guarantees}.

Given that sentences must be concatenated to form a document (see~\autoref{tab:table-data}), the document-level datasets are necessarily smaller than the original ones. Consequently, the $\sigma$ noise introduced during document-level training with the DP-SGD algorithm is increased and a higher sampling rate is employed, in order to match the exact same $\varepsilon$ value among the two configurations. Apart from this, the DP-SGD hyperparameters, such as the gradient clip value $C$, are identical for both settings. Furthermore, we use more epochs to train document-level model in private setting to obtain decent performance (see~\autoref{sec:hyperparameters}).

\begin{figure*}[!t]
\centering
\begin{subfigure}{.33\textwidth}
  \centering
  \includegraphics[width=\linewidth]{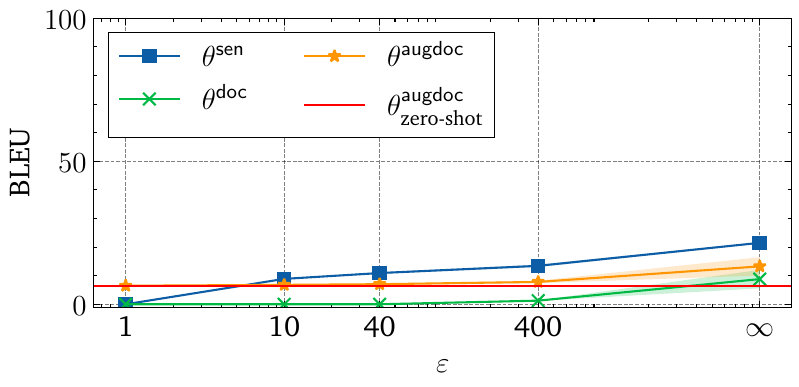}
  \caption{BSD}
  \label{fig:bleu-sen-test-bsd}
\end{subfigure}
\begin{subfigure}{.315\textwidth}
  \centering
  \includegraphics[width=\linewidth]{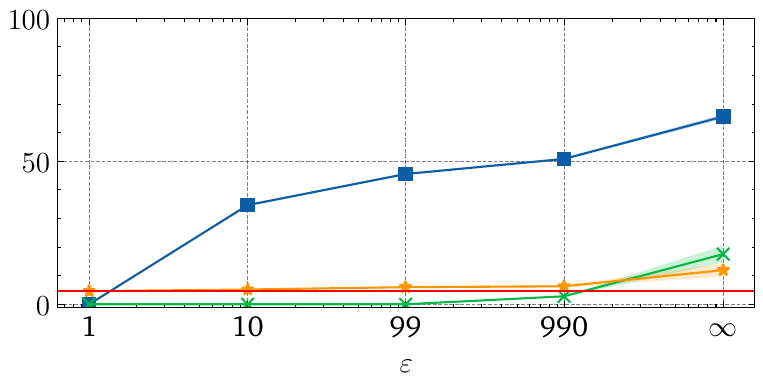}
  \caption{MAIA}
  \label{fig:bleu-sen-test-maia}
\end{subfigure}%
\begin{subfigure}{.315\textwidth}
  \centering
  \includegraphics[width=\linewidth]{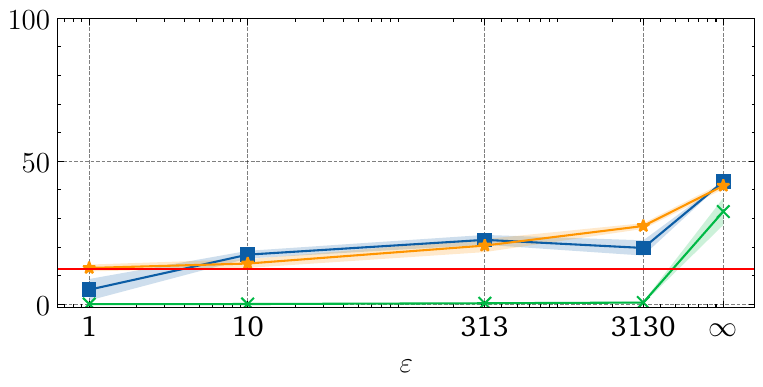}
  \caption{Europarl}
  \label{fig:bleu-sen-test-europarl}
\end{subfigure}%
\caption{BLEU scores on \dtestsen~for all four model fine-tuning configurations. Lower $\varepsilon$ corresponds to better privacy.}
\label{fig:bleu-sen-test}
\end{figure*}

\subsection{Evaluation} \label{eval:performance}

\paragraph{Performance}
We report BLEU~\cite{papineni2002bleu} for $n$-gram matching evaluation and BERTScore~\citep{zhang2019bertscore} for semantic similarity evaluation. We refer to \autoref{bertscore-modification} for the details of our modification to BERTScore for evaluation on long texts.

\paragraph{MIA}
It is difficult to know whether a sentence belongs to the training data of a document-level model or a sentence-level model.
For each dataset at sentence level, we consider the validation set and the test set as \emph{non-members} and the training set as \emph{members}.
However, the training set is huge compared to the validation set and the test set; it is recommended to balance the dataset for MIA evaluation to avoid the bias of the attacker~\citep{jayaraman-2019-evaluating}.
We sample the total number of members from the training set to be equal to the total number of examples in the validation set and the test set for our experiments, similar to~\citet{yeom-2018, jayaraman-2019-evaluating}.
Formally, let $\alpha$ be the total number of sentences in the validation set $\mathcal{D}_{\mathrm{val}}^{\mathsf{sen}}$ and the test set $\mathcal{D}_{\mathrm{test}}^{\mathsf{sen}}$.
By leveraging Sampling Without Replacement to avoid duplicating instances, we have a sampled set of sentences $\mathcal{D}_{\mathrm{sampled}}^{\mathsf{sen}}$ from a sentence-level training set $\mathcal{D}_{\mathrm{train}}^{\mathsf{sen}}$:
$(s_i, r_i),\dots, (s_\alpha, r_\alpha)\sim\mathcal{D}_{\mathrm{train}}^\mathtt{sen}$,
where $s_i$ is the source and $r_i$ is the corresponding target sentence for~$i\in\{1,\dots,\alpha\}$ and $|\mathcal{D}_{\mathrm{val}}^{\mathsf{sampled}}| = |$\dvalsen$| + |$\dtestsen$|$ (See \autoref{tab:my-table-mia} in \autoref{sec:stats-attack} for the exact number).

\paragraph{PII}
In NMT, the cross-lingual PII detection might be not comparable between languages of input $s$ and target $r$.
Hence, we only report the PII leakage estimation on the target language based on the reference $r$
(See Table~\ref{tab:my-table-pii} in Appendix for the number of PII in $\mathcal{D}_{\mathrm{sampled}}^{\mathsf{sen}}$ of each dataset).
We use the percentage of PII leakage estimation as the metric, since all PIIs are extracted from the sampled training data.
Namely, this is the ratio of the number of detected PII data to the total number of PII data in the sampled training data:
\begin{equation}
    \text{PII}_{\text{\% leakage}}(\mathbf{r})=\dfrac{\text{PII of TP }\mathbf{r} \in\mathcal{D}_{\mathrm{sampled}}^{\mathsf{sen}}}{\text{Total PII of }\mathbf{r} \in \mathcal{D}_{\mathrm{sampled}}^{\mathsf{sen}}}
\end{equation}
We suspect a private training model to have a lower PII leakage percentage than 50\%.

\section{Results}\label{sec:results}

\subsection{Privacy/utility trade-off}
We present the BLEU score resutls below, we refer to Appendix~\ref{sec:appx-detailed-results} for BERTScore results.

\paragraph{Evaluation on \dtestdoc}

Figure~\ref{fig:bleu-doc-test} shows the BLEU score of the two approaches on~\dtestdoc.
As expected, we can observe the deterioration of the BLEU score as the value of~$\varepsilon$ decreases on both datasets.
\textbf{The additional training data from WMT22 is beneficial for the translation quality on both datasets}.
Without pre-training on WMT22, the BLEU score of~\docm~is significantly lower than~\augm, with in a significant drop in translation quality.
Moreover, the BLEU score of~\zaugm, which is fine-tuned on WMT22, is even higher than the fine-tuned~\docm~at $\varepsilon=\infty$ on the BSD dataset.
The results of~\augm~are consistently better than~\docm~across all values of~$\varepsilon$.
Overall, these results indicate that privately fine-tuning on the target task is slightly beneficial for the translation quality, with respect to the domain translation quality at the document level.

\paragraph{Evaluation of \dtestsen}

Figure~\ref{fig:bleu-sen-test} shows the BLEU score of the three approaches on~\dtestsen.
On MAIA, the BLEU score of~\senm~is superior to~\docm~and~\augm~at any chosen value of~$\varepsilon$, except for $\varepsilon=1$.
Even at $\varepsilon=10$, the BLEU score of~\senm~is very high at 35 BLEU score, while the BLEU score of all document-level models are low.

On BSD, the performance gap between~\senm,~\docm~and~\augm~is less significant, possibly due to the distantly related language pair.
The difference in BLEU score between~\senm~and~\augm~is about 10 for $\varepsilon > 10$.
Pre-training on WMT22 is beneficial for the translation quality at the sentence level on the BSD dataset, since the BLEU score of~\zaugm~is already higher than~\docm~at $\varepsilon=\infty$.

On Europarl, we also observe that the results of~\senm,~\docm~and~\augm~at $\varepsilon=\infty$ are very close. Similar to the previous results on \dtestdoc, we could not achieve any sort of performance level when training~\docm~with DP-SGD.

We refer to \autoref{tab:dtestsen-maia-sample}, \autoref{tab:dtestsen-BSD-sample} and \autoref{tab:dtestsen-Europarl-sample}
in Appendix~\ref{sec:more_discussion} for specific translation examples of each dataset.

\subsection{Privacy risk evaluation}

\begin{figure}[!t]
\centering
\begin{subfigure}{\columnwidth}
  \centering
  \includegraphics[width=\linewidth]{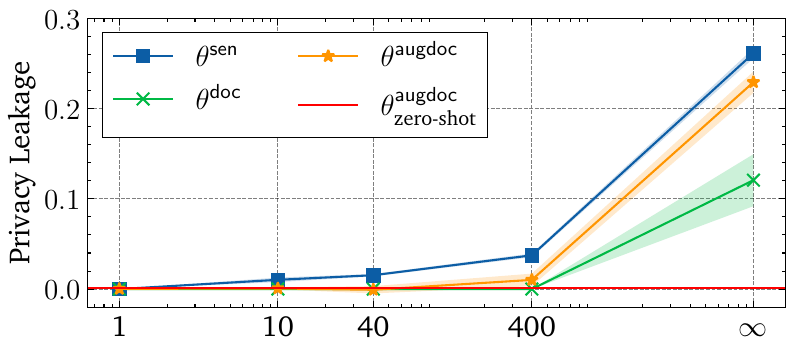}
  \caption{BSD}
  \label{fig:privacy-leakage-bsd}
\end{subfigure}
\hfill
\begin{subfigure}{\columnwidth}
  \centering
  \includegraphics[width=\linewidth]{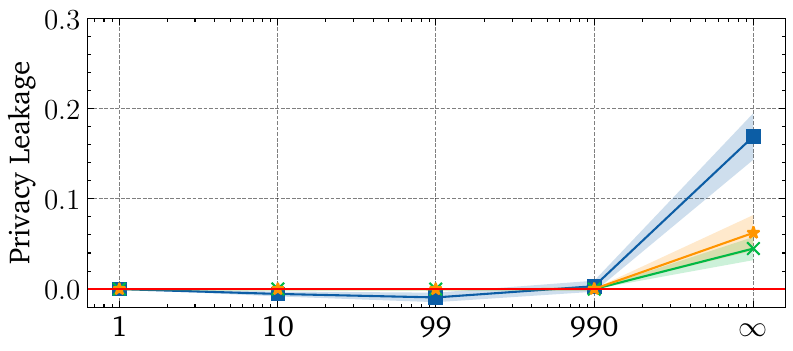}
  \caption{MAIA}
  \label{fig:privacy-leakage-maia}
\end{subfigure}%
\hfill
\begin{subfigure}{\columnwidth}
  \centering
  \includegraphics[width=\linewidth]{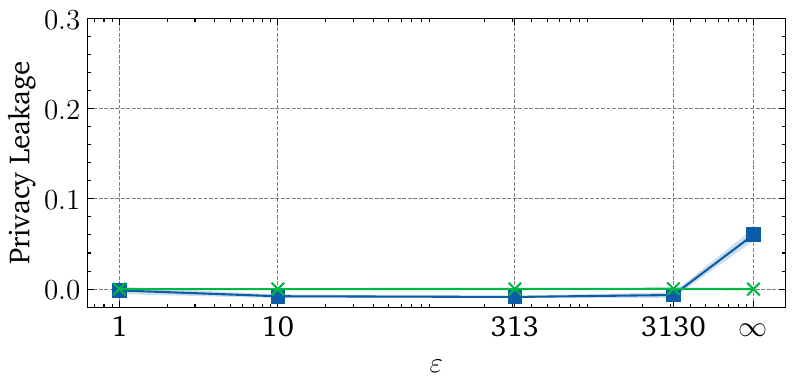}
  \caption{Europarl}
  \label{fig:privacy-leakage-europarl}
\end{subfigure}%
\caption{Privacy leakage using the loss-based MIA on $\mathcal{D}_{\mathrm{sampled}}^{\mathsf{sen}}$,~\dvalsen~and~\dtestsen, for all four model fine-tuning configurations.}
\label{fig:privacy-leakage}
\end{figure}

As in our assumption, we consider the adversary 
knows the average loss of the model on the training data, more particularly, the average loss of~\senm~on~\dtrainsen~at $\varepsilon=\infty$ before performing any attack evaluation.
\autoref{fig:privacy-leakage} shows the privacy leakage for both MAIA and BSD using loss-based MIA. 

\paragraph{MAIA}

On MAIA, 
\textbf{the privacy leakage via loss-based MIA on document-level training is lower than sentence-level training}.
Empirically, the leakage on \docm~and~\augm~is 50\% lower than~\senm~at $\varepsilon=\infty$.
Furthermore, the value continues to decrease and eventually drops below zero as the variable $\varepsilon$ decreases as FPR is higher than TPR, results in ineffective MIA.
The same applies to the other models,~\docm,~\augm, we observe no privacy leakage after applying DP due to the difficulty in optimizing longer sequences plus the noise added to the gradients according to the privacy budget.

\paragraph{BSD}
On BSD, \textbf{the attacker has a higher advantage 
at any level of} $\varepsilon$.
This implies that the training data has distinct characteristics that make it easier for the adversary to infer the membership of the training data vs.\ test data compared to MAIA\@.
At $\varepsilon=\infty$, the privacy leakage of~\augm~is only 0.03 points behind~\senm.
The privacy leakage on~\docm~\textbf{is also lower than}~\senm; however,~\augm~still has a small leakage at~$\varepsilon=400$~and the leakage of~\augm~converges to near zero at $\varepsilon=10$.

\paragraph{Europarl}
On Europarl,\footnote{We do not conduct the privacy risk and PII disclosure experiments with~\augm, since Europarl is part of the WMT dataset.} we find that the attacker has only a small advantage in normal training with~\senm, which may be due to the similar optimization for the domain of large training data, while~\docm~shows no leakage. Using DP-SGD, all models show no leakage as well. 

\subsection{PII disclosure} \label{subsec:pii-disclosure}
Despite DP mitigating the privacy leakage to a large extent, the true positive prediction from MIA after applying DP still plays a significant role in the privacy risk evaluation.
This helps us determine \textbf{the extent to which the model could potentially reveal PII} (issue of overfitting).

\paragraph{MAIA}
The PII leakage percentage of~\docm~and~\augm~is 0 after applying differential privacy.
At $\varepsilon=\infty$, the PII leakage percentage of those approaches is below 0.25, while \senm~is approximately 0.80 at $\varepsilon=\infty$ and down to 0.40 at $\varepsilon=10$.

\paragraph{BSD}
On BSD, surprisingly, the PII leakage percentage of~\augm~at $\varepsilon=\infty$ is slightly higher than~\senm.
Both are around 0.75, while \docm~is approximately 0.4.
The most interesting observation is that the PII leakage percentage of~\augm~deteriorates faster as $\varepsilon$ increases compared to~\senm, though it is higher at $\varepsilon=\infty$.
This is a significant sign that the~\docm~and~\augm~is more effective in reducing privacy risks than~\senm.

\paragraph{Europarl}
The results of PII leakage percentage on Europarl are similar to BSD and MAIA. \docm~is 0 after training with DP-SGD, while~\senm~leaks progressively less at lower $\varepsilon$ values. At $\varepsilon=1$, instead of being 0 as on MAIA and BSD, there is a small amount of PII leakage on Europarl with~\senm. However, the result is insignificant for the 50\% threshold of a private training model.

\section{Discussion}
\label{sec:discussion}

\subsection{Trade-off for privacy granularity}
Overall on \dtestdoc, the BLEU scores of~\augm~greatly vary across different values of~$\varepsilon$ in the private training setup on both datasets.
This results in higher mean and standard deviation of BLEU scores for~\augm~at the lower values of~$\varepsilon$, compared to higher values of~$\varepsilon$.
This may be due to the private training process being unstable when optimizing the objective function with respect to long sequences, which is the case of~\docm~in normal training.

For the results on \dtestsen
, the translation quality of~\senm~is superior to~\docm~and~\augm~at any chosen value~of~$\varepsilon > 1$ on both datasets.
Without additional training data from WMT22, the private training process of~\docm~again becomes unstable for long sequences. 
This instability results in a divergent loss in the training process and a significant drop in translation quality, even worse with the added noise from DP-SGD. 
As the base of~\augm,~\zaugm~benefits~\augm, which ensures the translation quality to be higher than zero in terms of BLEU 
even at $\varepsilon=1$ on both datasets.
The document-level training model's results also vary significantly across different values of $\varepsilon$ in the private training setups for both datasets.

The privacy/utility trade-off evaluation also shows that the language pair of the dataset has 
impact on the translation quality of the private training model.
To close the gap between the translation quality of~\senm~and~\augm~on \dtestsen
, training with more data should be considered.

\subsection{PII extraction and error analysis}

\autoref{tab:leakage-example} shows an example of leakage over three different $\varepsilon$ values across three model settings. 
Overall, without training with DP-SGD, we observe the model to overfit to utterances with PII, even when using~\docm~and~\augm. 
By training with DP-SGD, there is no leakage with the document-level models.
However, this is not the case for~\senm, even going down to $\varepsilon=10$, where the customer's name and website URL seem to be memorized.
This suggests that \senm~might require a more stringent privacy budget (lower $\varepsilon$) to prevent overfitting to sensitive information compared to document-level models.

\begin{figure}[!t]
\centering
\begin{subfigure}{\columnwidth}
  \centering
  \includegraphics[width=\linewidth]{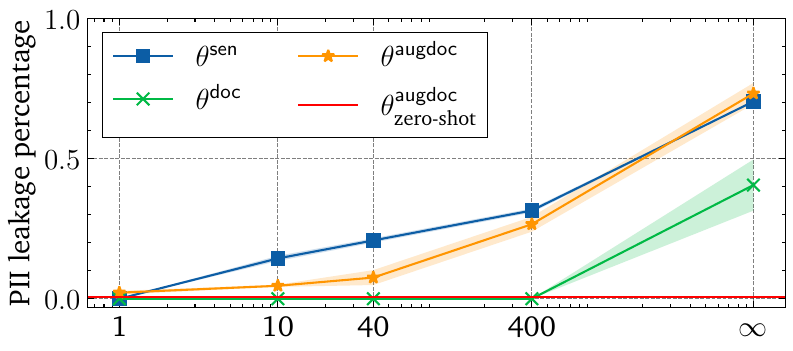}
  \caption{BSD}
  \label{fig:pII-guess-accuracy-bsd}
\end{subfigure}
\hfill
\begin{subfigure}{\columnwidth}
  \centering
  \includegraphics[width=\linewidth]{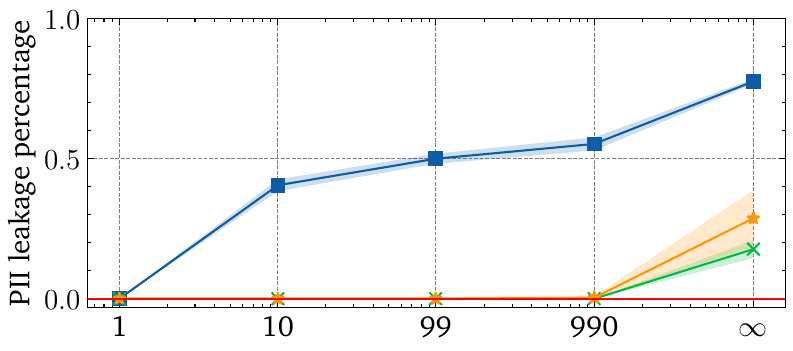}
  \caption{MAIA}
  \label{fig:pII-guess-accuracy-maia}
\end{subfigure}%
\hfill
\begin{subfigure}{\columnwidth}
  \centering
  \includegraphics[width=\linewidth]{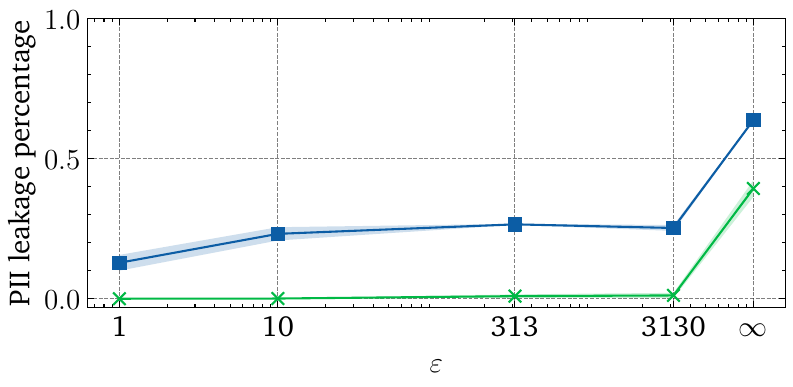}
  \caption{Europarl}
  \label{fig:pII-guess-accuracy-europarl}
\end{subfigure}%
\caption{PII leakage percentage on $\mathcal{D}_{\mathrm{sampled}}^{\mathsf{sen}}$.}
\label{fig:pII-guess-accuracy}
\end{figure}

\begin{table*}[tb]
\centering
\resizebox{\textwidth}{!}{%
\begin{tabular}{@{}lllll@{}}
\toprule
\textbf{Example}                                                                                   & \textbf{Epsilon} & \senm & \docm & \augm \\ \midrule
\multirow{5}{*}{Agent: Good Morning \colorbox{cyan}{Dipl.-Ing. Bastian Heuser}}                            & $\infty$& \checkmark   & \checkmark   & \checkmark      \\
                                                                                          & 990     & \checkmark   & $\times$   & $\times$      \\
                                                                                          & 10      & \checkmark   & $\times$   & $\times$      \\\midrule
\multirow{5}{*}{Agent: So you would like to cancel the reorder of the lamp?}              & $\infty$& \checkmark   & $\times$   & $\times$      \\
                                                                                          & 990     & \checkmark   & $\times$   & $\times$      \\
                                                                                          & 10      & $\times$   & $\times$   & $\times$      \\\midrule
\multirow{5}{*}{Customer: The standing lamp or the hanging lamp?}                         & $\infty$& \checkmark   & $\times$   & $\times$      \\
                                                                                          & 990     & \checkmark   & $\times$   & $\times$      \\
                                                                                          & 10      & \checkmark   & $\times$   & $\times$      \\\midrule
\multirow{5}{*}{Agent: - Go to \colorbox{cyan}{http://www.suessebier.de/}}                                 & $\infty$& \checkmark   & \checkmark   & \checkmark      \\
                                                                                          & 990     & \checkmark   & $\times$   & $\times$      \\
                                                                                          & 10      & \checkmark   & $\times$   & $\times$      \\ \midrule
\multirow{5}{*}{Agent: Do you have the order number that starts \colorbox{cyan}{160$\dots$..}?}            & $\infty$& \checkmark   & \checkmark   & \checkmark      \\
                                                                                          & 990     & \checkmark   & $\times$   & $\times$      \\
                                                                                          & 10      & $\times$   & $\times$   & $\times$      \\ \midrule
\multirow{5}{*}{Agent: Thank you - so the \colorbox{cyan}{Bärer GmbH} (140 x 200 cm),  \colorbox{cyan}{Samt} in  \colorbox{cyan}{Nachtgrau}?}  & $\infty$& \checkmark   & \checkmark   & \checkmark      \\
                                                                                          & 990     & \checkmark   & $\times$   & $\times$      \\
                                                                                          & 10      & $\times$   & $\times$   & $\times$      \\\bottomrule
\end{tabular}%
}
\caption{Examples of leakage from MAIA \dtrainsen~using the MIA and PII evaluation on sentence-level and document-level models. The utterances are collected from within a dialogue. The color depicts the selected PII by \texttt{Presidio}. \checkmark denotes leakage. $\times$ denotes no leakage.}
\label{tab:leakage-example}
\end{table*}

\section{Conclusion}
\label{sec:conclusion}

We have examined 
the privacy leakage of sentence-level vs.\ document-level approaches using the loss-based membership inference attack on the mLongT5 model.
The results show that a sentence-level model has more risks of privacy leakage than a document-level model.
Specifically, a sentence-level model is more likely to overfit compared to a document-level model, which can lead to more confident guessing of sentence-level training instances.

Furthermore, regarding the privacy/utility trade-off in the document-level model, optimizing transformer models for long texts, especially with DP-SGD, is a challenging task that requires more data than our downstream dataset.
We demonstrate our solution to this problem with an augmented training technique, using a large public dataset, in our case WMT22, to achieve an acceptable utility, then fine-tune on the downstream dataset to achieve the best privacy/utility trade-off.
For future work, we aim to design a better MIA that takes into account the correlation aspect of NLP datasets.

\section*{Limitations}

Although scaling DP to the document level in NLP shows promise, there are several notable pitfalls in our work that need to be addressed in future work:
  \begin{enumerate}[topsep=5pt,itemsep=0pt,leftmargin=*]
    \item The training data for the document-level scenario is insufficient. \label{2}
    \item Loss-based MIA does not consider the correlation between data and the PII evaluation schema is not perfect. \label{3and4}
  \end{enumerate}

For \autoref{2}, since we are aiming to find a sensitive dataset (e.g. multiple instances of PII), we are limited by the availability of such data for NMT.
We are also limited by the size of the document-level dataset when concatenating sentences, which is crucial for training a document-level model.
As~\citet{zhuocheng-etal-2023-scaling} suggest, we need at least four million training instances to outperform the sentence-level model, while even with the larger Europarl dataset we have around 140k.

Regarding \autoref{3and4}, the loss-based MIA has a significant limitation as it does not take into account the correlation between data.
We argue that document-level privacy is stronger than sentence-level privacy when considering this correlation.
Therefore, a better MIA method that considers the correlation between data is needed to prove this claim.
Future work should investigate the correlation between data and how it affects the privacy guarantee, such as~\citet{humphries_2023}, but for NLP tasks.
In addition, the PII evaluation schema focuses more on risk assessment than strict evaluation.
Our PII evaluation relies on the model's confidence in identifying true positive predictions from the MIA and the detection of PII is carried out automatically with \texttt{Presidio}, which may result in false positives.
Additionally, for the case of NMT, we only consider PII in the target language, assuming the attacker has access to both the source and target instances.
This is relevant to the performance of PII detection in source inputs which are not English.

\section*{Acknowledgements}

This project was supported by the PrivaLingo research grant (Hessisches Ministerium des Innern und für Sport).
This work has also been partly supported by the Research Center Trustworthy Data
Science and Security (\url{https://rc-trust.ai}), one of the Research Alliance
centers within the UA Ruhr (\url{https://uaruhr.de}).
Thanks to Erion Çano and Lena Held for helpful feedback on a preliminary draft.

\bibliography{bibliography, need2format}

\begin{thebibliography}{47}
\expandafter\ifx\csname natexlab\endcsname\relax\def\natexlab#1{#1}\fi

\bibitem[{Abadi et~al.(2016)Abadi, Chu, Goodfellow, McMahan, Mironov, Talwar, and Zhang}]{abadi2016deep}
Martin Abadi, Andy Chu, Ian Goodfellow, H~Brendan McMahan, Ilya Mironov, Kunal Talwar, and Li~Zhang. 2016.
\newblock Deep learning with differential privacy.
\newblock In \emph{Proceedings of the 2016 ACM SIGSAC conference on computer and communications security}, pages 308--318.

\bibitem[{Anil et~al.(2022)Anil, Ghazi, Gupta, Kumar, and Manurangsi}]{anil-etal-2022-large}
Rohan Anil, Badih Ghazi, Vineet Gupta, Ravi Kumar, and Pasin Manurangsi. 2022.
\newblock \href {https://aclanthology.org/2022.findings-emnlp.484} {Large-scale differentially private {BERT}}.
\newblock In \emph{Findings of the Association for Computational Linguistics: EMNLP 2022}, pages 6481--6491, Abu Dhabi, United Arab Emirates. Association for Computational Linguistics.

\bibitem[{Beltagy et~al.(2020)Beltagy, Peters, and Cohan}]{beltagy2020longformer}
Iz~Beltagy, Matthew~E Peters, and Arman Cohan. 2020.
\newblock Longformer: The long-document transformer.
\newblock \emph{arXiv preprint arXiv:2004.05150}.

\bibitem[{Bradbury et~al.(2018)Bradbury, Frostig, Hawkins, Johnson, Leary, Maclaurin, Necula, Paszke, Vander{P}las, Wanderman-{M}ilne, and Zhang}]{jax2018github}
James Bradbury, Roy Frostig, Peter Hawkins, Matthew~James Johnson, Chris Leary, Dougal Maclaurin, George Necula, Adam Paszke, Jake Vander{P}las, Skye Wanderman-{M}ilne, and Qiao Zhang. 2018.
\newblock \href {http://github.com/google/jax} {{JAX}: composable transformations of {P}ython+{N}um{P}y programs}.
\newblock \texttt{http://github.com/google/jax}.

\bibitem[{Brown et~al.(2022)Brown, Lee, Mireshghallah, Shokri, and Tram{\`e}r}]{brown2022does}
Hannah Brown, Katherine Lee, Fatemehsadat Mireshghallah, Reza Shokri, and Florian Tram{\`e}r. 2022.
\newblock What does it mean for a language model to preserve privacy?
\newblock In \emph{Proceedings of the 2022 ACM Conference on Fairness, Accountability, and Transparency}, pages 2280--2292.

\bibitem[{Carlini et~al.(2021)Carlini, Tramer, Wallace, Jagielski, Herbert-Voss, Lee, Roberts, Brown, Song, Erlingsson et~al.}]{carlini2021extracting}
Nicholas Carlini, Florian Tramer, Eric Wallace, Matthew Jagielski, Ariel Herbert-Voss, Katherine Lee, Adam Roberts, Tom Brown, Dawn Song, Ulfar Erlingsson, et~al. 2021.
\newblock Extracting training data from large language models.
\newblock In \emph{30th USENIX Security Symposium (USENIX Security 21)}, pages 2633--2650.

\bibitem[{Dwork and Roth(2013)}]{Dwork.Roth.2013}
Cynthia Dwork and Aaron Roth. 2013.
\newblock \href {https://doi.org/10.1561/0400000042} {{The Algorithmic Foundations of Differential Privacy}}.
\newblock \emph{Foundations and Trends{\textregistered} in Theoretical Computer Science}, 9(3-4):211--407.

\bibitem[{Farinha et~al.(2022)Farinha, Farajian, Buchicchio, Fernandes, C.~de Souza, Moniz, and Martins}]{farinha-etal-2022-findings}
Ana~C Farinha, M.~Amin Farajian, Marianna Buchicchio, Patrick Fernandes, Jos{\'e}~G. C.~de Souza, Helena Moniz, and Andr{\'e} F.~T. Martins. 2022.
\newblock \href {https://aclanthology.org/2022.wmt-1.70} {Findings of the {WMT} 2022 shared task on chat translation}.
\newblock In \emph{Proceedings of the Seventh Conference on Machine Translation (WMT)}, pages 724--743, Abu Dhabi, United Arab Emirates (Hybrid). Association for Computational Linguistics.

\bibitem[{Fernandes et~al.(2021)Fernandes, Yin, Neubig, and Martins}]{fernandes_measuring_2021}
Patrick Fernandes, Kayo Yin, Graham Neubig, and André F.~T. Martins. 2021.
\newblock \href {https://doi.org/10.18653/v1/2021.acl-long.505} {Measuring and {Increasing} {Context} {Usage} in {Context}-{Aware} {Machine} {Translation}}.
\newblock In \emph{Proceedings of the 59th {Annual} {Meeting} of the {Association} for {Computational} {Linguistics} and the 11th {International} {Joint} {Conference} on {Natural} {Language} {Processing} ({Volume} 1: {Long} {Papers})}, pages 6467--6478, Online. Association for Computational Linguistics.

\bibitem[{Guo et~al.(2022)Guo, Ainslie, Uthus, Ontanon, Ni, Sung, and Yang}]{guo_longt5_2022}
Mandy Guo, Joshua Ainslie, David Uthus, Santiago Ontanon, Jianmo Ni, Yun-Hsuan Sung, and Yinfei Yang. 2022.
\newblock \href {https://doi.org/10.18653/v1/2022.findings-naacl.55} {{LongT5}: {Efficient} {Text}-{To}-{Text} {Transformer} for {Long} {Sequences}}.
\newblock In \emph{Findings of the {Association} for {Computational} {Linguistics}: {NAACL} 2022}, pages 724--736, Seattle, United States. Association for Computational Linguistics.

\bibitem[{Heek et~al.(2023)Heek, Levskaya, Oliver, Ritter, Rondepierre, Steiner, and van {Z}ee}]{flax2020github}
Jonathan Heek, Anselm Levskaya, Avital Oliver, Marvin Ritter, Bertrand Rondepierre, Andreas Steiner, and Marc van {Z}ee. 2023.
\newblock \href {http://github.com/google/flax} {{F}lax: A neural network library and ecosystem for {JAX}}.
\newblock \texttt{http://github.com/google/flax}.

\bibitem[{Hendy et~al.(2023)Hendy, Abdelrehim, Sharaf, Raunak, Gabr, Matsushita, Kim, Afify, and Awadalla}]{hendy_how_2023}
Amr Hendy, Mohamed Abdelrehim, Amr Sharaf, Vikas Raunak, Mohamed Gabr, Hitokazu Matsushita, Young~Jin Kim, Mohamed Afify, and Hany~Hassan Awadalla. 2023.
\newblock \href {http://arxiv.org/abs/2302.09210} {How {Good} {Are} {GPT} {Models} at {Machine} {Translation}? {A} {Comprehensive} {Evaluation}}.
\newblock ArXiv:2302.09210 [cs].

\bibitem[{Hisamoto et~al.(2020)Hisamoto, Post, and Duh}]{hisamoto2020membership}
Sorami Hisamoto, Matt Post, and Kevin Duh. 2020.
\newblock Membership inference attacks on sequence-to-sequence models: Is my data in your machine translation system?
\newblock \emph{Transactions of the Association for Computational Linguistics}, 8:49--63.

\bibitem[{Hu et~al.(2022)Hu, Salcic, Sun, Dobbie, Yu, and Zhang}]{hu_membership_2022}
Hongsheng Hu, Zoran Salcic, Lichao Sun, Gillian Dobbie, Philip~S. Yu, and Xuyun Zhang. 2022.
\newblock \href {https://doi.org/10.1145/3523273} {Membership {Inference} {Attacks} on {Machine} {Learning}: {A} {Survey}}.
\newblock \emph{ACM Computing Surveys}, 54(11s):1--37.

\bibitem[{Hu et~al.(2024)Hu, Habernal, Shen, and Wang}]{hu-etal-2024-differentially}
Lijie Hu, Ivan Habernal, Lei Shen, and Di~Wang. 2024.
\newblock \href {https://aclanthology.org/2024.findings-eacl.33} {{Differentially Private Natural Language Models: Recent Advances and Future Directions}}.
\newblock In \emph{Findings of the Association for Computational Linguistics: EACL 2024}, pages 478--499, St. Julian{'}s, Malta. Association for Computational Linguistics.

\bibitem[{Humphries et~al.(2023)Humphries, Oya, Tulloch, Rafuse, Goldberg, Hengartner, and Kerschbaum}]{humphries_2023}
Thomas Humphries, Simon Oya, Lindsey Tulloch, Matthew Rafuse, Ian Goldberg, Urs Hengartner, and Florian Kerschbaum. 2023.
\newblock \href {https://doi.org/10.1109/CSF57540.2023.00013} {{Investigating Membership Inference Attacks under Data Dependencies}}.
\newblock In \emph{2023 IEEE 36th Computer Security Foundations Symposium (CSF)}, pages 473--488, Los Alamitos, CA, USA. IEEE Computer Society.

\bibitem[{Igamberdiev and Habernal(2023)}]{igamberdiev-habernal-2023-dp}
Timour Igamberdiev and Ivan Habernal. 2023.
\newblock \href {https://doi.org/10.18653/v1/2023.findings-acl.874} {{DP}-{BART} for privatized text rewriting under local differential privacy}.
\newblock In \emph{Findings of the Association for Computational Linguistics: ACL 2023}, pages 13914--13934, Toronto, Canada. Association for Computational Linguistics.

\bibitem[{Igamberdiev et~al.(2024)Igamberdiev, Vu, Kuennecke, Yu, Holmer, and Habernal}]{igamberdiev-etal-2024-dp}
Timour Igamberdiev, Doan Nam~Long Vu, Felix Kuennecke, Zhuo Yu, Jannik Holmer, and Ivan Habernal. 2024.
\newblock \href {https://aclanthology.org/2024.eacl-demo.11} {{DP}-{NMT}: Scalable differentially private machine translation}.
\newblock In \emph{Proceedings of the 18th Conference of the European Chapter of the Association for Computational Linguistics: System Demonstrations}, pages 94--105, St. Julians, Malta. Association for Computational Linguistics.

\bibitem[{Jayaraman and Evans(2019)}]{jayaraman-2019-evaluating}
Bargav Jayaraman and David Evans. 2019.
\newblock \href {https://www.usenix.org/conference/usenixsecurity19/presentation/jayaraman} {Evaluating differentially private machine learning in practice}.
\newblock In \emph{28th USENIX Security Symposium (USENIX Security 19)}, pages 1895--1912, Santa Clara, CA. USENIX Association.

\bibitem[{Karpinska and Iyyer(2023)}]{karpinska-iyyer-2023-large}
Marzena Karpinska and Mohit Iyyer. 2023.
\newblock \href {https://doi.org/10.18653/v1/2023.wmt-1.41} {Large language models effectively leverage document-level context for literary translation, but critical errors persist}.
\newblock In \emph{Proceedings of the Eighth Conference on Machine Translation}, pages 419--451, Singapore. Association for Computational Linguistics.

\bibitem[{Kocmi et~al.(2022)Kocmi, Bawden, Bojar, Dvorkovich, Federmann, Fishel, Gowda, Graham, Grundkiewicz, Haddow, Knowles, Koehn, Monz, Morishita, Nagata, Nakazawa, Nov{\'a}k, Popel, and Popovi{\'c}}]{kocmi-etal-2022-findings}
Tom Kocmi, Rachel Bawden, Ond{\v{r}}ej Bojar, Anton Dvorkovich, Christian Federmann, Mark Fishel, Thamme Gowda, Yvette Graham, Roman Grundkiewicz, Barry Haddow, Rebecca Knowles, Philipp Koehn, Christof Monz, Makoto Morishita, Masaaki Nagata, Toshiaki Nakazawa, Michal Nov{\'a}k, Martin Popel, and Maja Popovi{\'c}. 2022.
\newblock \href {https://aclanthology.org/2022.wmt-1.1} {Findings of the 2022 conference on machine translation ({WMT}22)}.
\newblock In \emph{Proceedings of the Seventh Conference on Machine Translation (WMT)}, pages 1--45, Abu Dhabi, United Arab Emirates (Hybrid). Association for Computational Linguistics.

\bibitem[{Koehn(2005)}]{koehn-2005-europarl}
Philipp Koehn. 2005.
\newblock \href {https://aclanthology.org/2005.mtsummit-papers.11} {{E}uroparl: A parallel corpus for statistical machine translation}.
\newblock In \emph{Proceedings of Machine Translation Summit X: Papers}, pages 79--86, Phuket, Thailand.

\bibitem[{Kudo and Richardson(2018)}]{kudo-richardson-2018-sentencepiece}
Taku Kudo and John Richardson. 2018.
\newblock \href {https://doi.org/10.18653/v1/D18-2012} {{S}entence{P}iece: A simple and language independent subword tokenizer and detokenizer for neural text processing}.
\newblock In \emph{Proceedings of the 2018 Conference on Empirical Methods in Natural Language Processing: System Demonstrations}, pages 66--71, Brussels, Belgium. Association for Computational Linguistics.

\bibitem[{Li et~al.(2022)Li, Tramer, Liang, and Hashimoto}]{li2022large}
Xuechen Li, Florian Tramer, Percy Liang, and Tatsunori Hashimoto. 2022.
\newblock \href {https://openreview.net/forum?id=bVuP3ltATMz} {Large language models can be strong differentially private learners}.
\newblock In \emph{International Conference on Learning Representations}.

\bibitem[{Liu et~al.(2020)Liu, Gu, Goyal, Li, Edunov, Ghazvininejad, Lewis, and Zettlemoyer}]{liu2020multilingual}
Yinhan Liu, Jiatao Gu, Naman Goyal, Xian Li, Sergey Edunov, Marjan Ghazvininejad, Mike Lewis, and Luke Zettlemoyer. 2020.
\newblock Multilingual denoising pre-training for neural machine translation.
\newblock \emph{Transactions of the Association for Computational Linguistics}, 8:726--742.

\bibitem[{Liu et~al.(2019)Liu, Ott, Goyal, Du, Joshi, Chen, Levy, Lewis, Zettlemoyer, and Stoyanov}]{liu2019roberta}
Yinhan Liu, Myle Ott, Naman Goyal, Jingfei Du, Mandar Joshi, Danqi Chen, Omer Levy, Mike Lewis, Luke Zettlemoyer, and Veselin Stoyanov. 2019.
\newblock Roberta: A robustly optimized bert pretraining approach.
\newblock \emph{arXiv preprint arXiv:1907.11692}.

\bibitem[{Mireshghallah et~al.(2022)Mireshghallah, Uniyal, Wang, Evans, and Berg-Kirkpatrick}]{mireshghallah-etal-2022-empirical}
Fatemehsadat Mireshghallah, Archit Uniyal, Tianhao Wang, David Evans, and Taylor Berg-Kirkpatrick. 2022.
\newblock \href {https://doi.org/10.18653/v1/2022.emnlp-main.119} {An empirical analysis of memorization in fine-tuned autoregressive language models}.
\newblock In \emph{Proceedings of the 2022 Conference on Empirical Methods in Natural Language Processing}, pages 1816--1826, Abu Dhabi, United Arab Emirates. Association for Computational Linguistics.

\bibitem[{Mironov(2017)}]{mironov_renyi_2017}
Ilya Mironov. 2017.
\newblock \href {https://doi.org/10.1109/CSF.2017.11} {Rényi {Differential} {Privacy}}.
\newblock In \emph{2017 {IEEE} 30th {Computer} {Security} {Foundations} {Symposium} ({CSF})}, pages 263--275, Santa Barbara, CA. IEEE.

\bibitem[{Moosavi et~al.(2021)Moosavi, R{\"u}ckl{\'e}, Roth, and Gurevych}]{moosavi2021scigen}
Nafise~Sadat Moosavi, Andreas R{\"u}ckl{\'e}, Dan Roth, and Iryna Gurevych. 2021.
\newblock \href {https://openreview.net/forum?id=Jul-uX7EV_I} {Scigen: a dataset for reasoning-aware text generation from scientific tables}.
\newblock In \emph{Thirty-fifth Conference on Neural Information Processing Systems Datasets and Benchmarks Track (Round 2)}.

\bibitem[{Papineni et~al.(2002)Papineni, Roukos, Ward, and Zhu}]{papineni2002bleu}
Kishore Papineni, Salim Roukos, Todd Ward, and Wei-Jing Zhu. 2002.
\newblock \href {https://doi.org/10.3115/1073083.1073135} {{B}leu: a method for automatic evaluation of machine translation}.
\newblock In \emph{Proceedings of the 40th Annual Meeting of the Association for Computational Linguistics}, pages 311--318, Philadelphia, Pennsylvania, USA. Association for Computational Linguistics.

\bibitem[{Ponomareva et~al.(2022)Ponomareva, Bastings, and Vassilvitskii}]{ponomareva-etal-2022-training}
Natalia Ponomareva, Jasmijn Bastings, and Sergei Vassilvitskii. 2022.
\newblock \href {https://doi.org/10.18653/v1/2022.findings-acl.171} {Training text-to-text transformers with privacy guarantees}.
\newblock In \emph{Findings of the Association for Computational Linguistics: ACL 2022}, pages 2182--2193, Dublin, Ireland. Association for Computational Linguistics.

\bibitem[{Ponomareva et~al.(2023)Ponomareva, Hazimeh, Kurakin, Xu, Denison, McMahan, Vassilvitskii, Chien, and Thakurta}]{ponomareva2023dp}
Natalia Ponomareva, Hussein Hazimeh, Alex Kurakin, Zheng Xu, Carson Denison, H~Brendan McMahan, Sergei Vassilvitskii, Steve Chien, and Abhradeep~Guha Thakurta. 2023.
\newblock How to dp-fy ml: A practical guide to machine learning with differential privacy.
\newblock \emph{Journal of Artificial Intelligence Research}, 77:1113--1201.

\bibitem[{Post and Junczys-Dowmunt(2023)}]{post_escaping_2023}
Matt Post and Marcin Junczys-Dowmunt. 2023.
\newblock \href {http://arxiv.org/abs/2304.12959} {Escaping the sentence-level paradigm in machine translation}.
\newblock ArXiv:2304.12959 [cs].

\bibitem[{Radford et~al.(2018)Radford, Narasimhan, Salimans, Sutskever et~al.}]{radford2018improving}
Alec Radford, Karthik Narasimhan, Tim Salimans, Ilya Sutskever, et~al. 2018.
\newblock \href {https://s3-us-west-2.amazonaws.com/openai-assets/research-covers/language-unsupervised/language_understanding_paper.pdf} {Improving language understanding by generative pre-training}.

\bibitem[{Raffel et~al.(2020)Raffel, Shazeer, Roberts, Lee, Narang, Matena, Zhou, Li, and Liu}]{colin-2020-t5}
Colin Raffel, Noam Shazeer, Adam Roberts, Katherine Lee, Sharan Narang, Michael Matena, Yanqi Zhou, Wei Li, and Peter~J. Liu. 2020.
\newblock \href {http://jmlr.org/papers/v21/20-074.html} {Exploring the limits of transfer learning with a unified text-to-text transformer}.
\newblock \emph{Journal of Machine Learning Research}, 21(140):1--67.

\bibitem[{Rikters et~al.(2019)Rikters, Ri, Li, and Nakazawa}]{rikters2019designing}
Mat{\=\i}ss Rikters, Ryokan Ri, Tong Li, and Toshiaki Nakazawa. 2019.
\newblock \href {https://doi.org/10.18653/v1/D19-5204} {Designing the business conversation corpus}.
\newblock In \emph{Proceedings of the 6th Workshop on Asian Translation}, pages 54--61, Hong Kong, China. Association for Computational Linguistics.

\bibitem[{Senge et~al.(2022)Senge, Igamberdiev, and Habernal}]{Senge.et.al.2022.EMNLP}
Manuel Senge, Timour Igamberdiev, and Ivan Habernal. 2022.
\newblock \href {https://doi.org/10.18653/v1/2022.emnlp-main.496} {{One size does not fit all: Investigating strategies for differentially-private learning across NLP tasks}}.
\newblock In \emph{Proceedings of the 2022 Conference on Empirical Methods in Natural Language Processing}, pages 7340--7353, Abu Dhabi, UAE.

\bibitem[{Shokri et~al.(2017)Shokri, Stronati, Song, and Shmatikov}]{shokri_membership_2017}
Reza Shokri, Marco Stronati, Congzheng Song, and Vitaly Shmatikov. 2017.
\newblock \href {https://doi.org/10.1109/SP.2017.41} {Membership {Inference} {Attacks} {Against} {Machine} {Learning} {Models}}.
\newblock In \emph{2017 {IEEE} {Symposium} on {Security} and {Privacy} ({SP})}, pages 3--18.
\newblock ISSN: 2375-1207.

\bibitem[{Uthus et~al.(2023)Uthus, Ontanon, Ainslie, and Guo}]{uthus_mlongt5_2023}
David Uthus, Santiago Ontanon, Joshua Ainslie, and Mandy Guo. 2023.
\newblock \href {https://aclanthology.org/2023.findings-emnlp.628} {m{L}ong{T}5: A multilingual and efficient text-to-text transformer for longer sequences}.
\newblock In \emph{Findings of the Association for Computational Linguistics: EMNLP 2023}, pages 9380--9386, Singapore. Association for Computational Linguistics.

\bibitem[{Wicks and Post(2023)}]{wicks_identifying_2023}
Rachel Wicks and Matt Post. 2023.
\newblock \href {https://aclanthology.org/2023.wmt-1.42} {Identifying context-dependent translations for evaluation set production}.
\newblock In \emph{Proceedings of the Eighth Conference on Machine Translation}, pages 452--467, Singapore. Association for Computational Linguistics.

\bibitem[{Wu et~al.(2023)Wu, Foster, Qu, and Haffari}]{wu_document_2023}
Minghao Wu, George Foster, Lizhen Qu, and Gholamreza Haffari. 2023.
\newblock \href {https://doi.org/10.18653/v1/2023.eacl-main.33} {Document {Flattening}: {Beyond} {Concatenating} {Context} for {Document}-{Level} {Neural} {Machine} {Translation}}.
\newblock In \emph{Proceedings of the 17th {Conference} of the {European} {Chapter} of the {Association} for {Computational} {Linguistics}}, pages 448--462, Dubrovnik, Croatia. Association for Computational Linguistics.

\bibitem[{Xue et~al.(2021)Xue, Constant, Roberts, Kale, Al-Rfou, Siddhant, Barua, and Raffel}]{xue-etal-2021-mt5}
Linting Xue, Noah Constant, Adam Roberts, Mihir Kale, Rami Al-Rfou, Aditya Siddhant, Aditya Barua, and Colin Raffel. 2021.
\newblock \href {https://doi.org/10.18653/v1/2021.naacl-main.41} {m{T}5: A massively multilingual pre-trained text-to-text transformer}.
\newblock In \emph{Proceedings of the 2021 Conference of the North American Chapter of the Association for Computational Linguistics: Human Language Technologies}, pages 483--498, Online. Association for Computational Linguistics.

\bibitem[{Yeom et~al.(2018)Yeom, Giacomelli, Fredrikson, and Jha}]{yeom-2018}
Samuel Yeom, Irene Giacomelli, Matt Fredrikson, and Somesh Jha. 2018.
\newblock \href {https://doi.org/10.1109/CSF.2018.00027} {Privacy risk in machine learning: Analyzing the connection to overfitting}.
\newblock In \emph{2018 IEEE 31st Computer Security Foundations Symposium (CSF)}, pages 268--282.

\bibitem[{Yin and Habernal(2022)}]{yin-habernal-2022-privacy}
Ying Yin and Ivan Habernal. 2022.
\newblock \href {https://doi.org/10.18653/v1/2022.nllp-1.14} {Privacy-preserving models for legal natural language processing}.
\newblock In \emph{Proceedings of the Natural Legal Language Processing Workshop 2022}, pages 172--183, Abu Dhabi, United Arab Emirates (Hybrid). Association for Computational Linguistics.

\bibitem[{Zhang et~al.(2020)Zhang, Kishore, Wu, Weinberger, and Artzi}]{zhang2019bertscore}
Tianyi Zhang, Varsha Kishore, Felix Wu, Kilian~Q Weinberger, and Yoav Artzi. 2020.
\newblock \href {https://openreview.net/forum?id=SkeHuCVFDr} {{BERTScore: Evaluating Text Generation with BERT}}.
\newblock In \emph{International Conference on Learning Representations}.

\bibitem[{Zhuocheng et~al.(2023{\natexlab{a}})Zhuocheng, Gu, Zhang, and Feng}]{zhuocheng-etal-2023-addressing}
Zhang Zhuocheng, Shuhao Gu, Min Zhang, and Yang Feng. 2023{\natexlab{a}}.
\newblock \href {https://doi.org/10.18653/v1/2023.findings-emnlp.773} {Addressing the length bias challenge in document-level neural machine translation}.
\newblock In \emph{Findings of the Association for Computational Linguistics: EMNLP 2023}, pages 11545--11556, Singapore. Association for Computational Linguistics.

\bibitem[{Zhuocheng et~al.(2023{\natexlab{b}})Zhuocheng, Gu, Zhang, and Feng}]{zhuocheng-etal-2023-scaling}
Zhang Zhuocheng, Shuhao Gu, Min Zhang, and Yang Feng. 2023{\natexlab{b}}.
\newblock \href {https://doi.org/10.18653/v1/2023.findings-emnlp.556} {Scaling law for document neural machine translation}.
\newblock In \emph{Findings of the Association for Computational Linguistics: EMNLP 2023}, pages 8290--8303, Singapore. Association for Computational Linguistics.

\end{thebibliography}
\bibliographystyle{acl_natbib}

\appendix

\section{Differential privacy and DP-SGD}\label{sec:dp}

Differential privacy (DP)~\citep{Dwork.Roth.2013} is a mathematical framework that ensures that the output of an analysis on a dataset remains unchanged within a specific threshold when any data point is added or removed from the dataset.
More formally, for a privacy budget $\varepsilon \geq 0, \delta \in [0, 1]$, a mechanism $\mathcal{M} \colon D^n \to \mathcal{R}^k$ is $(\varepsilon, \delta)$ differentially private if for all datasets $D$ and $D'$ that differ in at most one instance, and for all $S \subseteq \mathrm{Range}(\mathcal{M})$:
\begin{equation}
    \mathrm{Pr}[\mathcal{M}(D) \in S] \leq \exp(\varepsilon) \cdot \mathrm{Pr}[\mathcal{M}(D') \in S] + \delta
\end{equation}
In other words, a mechanism $\mathcal{M}$ is $(\varepsilon, \delta)$ differentially private if the probability that the mechanism $\mathcal{M}$ returns a response $s \in S$ on dataset $D$ is at most $\exp(\varepsilon)$ times the probability of the mechanism $\mathcal{M}$ returns a response $s \in S$ on dataset $D'$.

According to the definition, the smaller the privacy budget $\varepsilon$, the greater privacy guarantees the $\mathcal{M}$ mechanism provides, due to its exponential nature, making the differing instance of $D$ and $D'$ indistinguishable.
This provides individuals with plausible deniability, as an attacker cannot be certain whether a specific instance belongs to the dataset $D$ or not.
However, choosing the appropriate privacy budget $\varepsilon$ is crucial to ensure the privacy guarantee of the $\mathcal{M}$ mechanism.
If the privacy budget is too large, the $\mathcal{M}$ mechanism will not provide a satisfactory privacy guarantee.
On the other hand, if the privacy budget is too low, then the noise added to the query is very high; thus making the mechanism $\mathcal{M}$ impractical.
To achieve the desired outcome, a compromise must be made between utility and privacy in the DP application, this is known as the \textit{privacy-utility trade-off}~\citep{Dwork.Roth.2013}.
Therefore, selecting the appropriate privacy budget for mechanism $\mathcal{M}$ is crucial.
In addition, the privacy budget is not fixed and can be adjusted.
The privacy budget is adjustable depending on the specific use case, data, and privacy preferences.
Typically, to achieve the $\mathcal{M}$ mechanism that satisfies the DP definition, we generally apply noise sampled from the Gaussian distribution to the `raw' output.

In the case of deep learning, we can apply DP during the training process, in particular prior to the optimization step of a deep neural network, acting as the data analyst, as in the case of DP-SGD~\citep{abadi2016deep}.
This method is presented in Algorithm~\ref{alg:dp_sgd}, in which the empirical loss function $L(\theta)$ is minimized with a noisy variant of SGD.
During each SGD step $t$, ~\cite{abadi2016deep} calculate the gradient $\nabla_{\theta} \mathcal{L}(\theta, x_i)$ for a random subset of samples $L_t$ via Poisson Sampling.
The $\ell_2$-norm of each gradient is then clipped, Gaussian noise $\mathcal{N}(0, \sigma^2 C^2 \mathbf{I})$ is added, and the average is taken over all noisy gradients for each element of $L_t$. 
A step is then taken in the reverse direction of this noisy gradient to update parameters $\theta_t$.
The algorithm aims to prevent over-optimization towards individual data points in the training dataset.

\begin{algorithm}[tb!]
    \SetAlgoNoLine
    \SetKwInput{Input}{Input}
    \SetKwInOut{Output}{Output}

    \Input{Examples $\{x_1,\dots, x_N\}$, loss function $\mathcal{L}(\theta)=\frac{1}{N}\sum_i \mathcal{L}(\theta, x_i)$. Parameters: learning rate $\eta_t$, noise scale $\sigma$, group size $L$, gradient norm bound $C$.}
    \textbf{Initialize} $\theta_0$ randomly\\
    \For{$t\in [T]$}{
        Take a random sample $L_t$ with sampling probability $L/N$\\
        \textbf{Compute gradient}\\
        For each $i \in L_t$, compute $\mathbf{g}_t(x_i) \leftarrow \nabla_{\theta_t} \mathcal{L}(\theta_t, x_i)$\\
        \textbf{Clip gradient}\\
        $\bar{\mathbf{g}}_t \leftarrow \mathbf{g}_t(x_i)/{\max\left(1, \frac{||\mathbf{g}_t(x_i)||_2}{C}\right)}$\\
        \textbf{Add noise}\\
        $\tilde{\mathbf{g}}_t \leftarrow \frac{1}{L}\left(\sum_i\overline{\mathbf{g}}_t(x_i) + \mathcal{N}(0, \sigma^2 C^2 \mathbf{I})\right)$\\
        \textbf{Descent}\\
        $\theta_{t+1} \leftarrow \theta_t - \eta_t \tilde{\mathbf{g}}_t$
    }
    \Output{$\theta_T$ and compute the overall privacy cost $(\varepsilon, \delta)$ using a privacy accounting method.}~\caption{Differentially private SGD (Outline) \citep{abadi2016deep}}
    \label{alg:dp_sgd}
\end{algorithm}

\section{Data preparation} \label{sec:data_prep}
\begin{figure*}[tb]
        \centering
                \begin{subfigure}{\linewidth}
            \centering
            \includegraphics{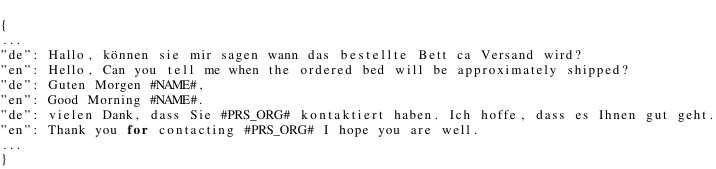}
         \caption{Sentence-level original training pairs}
     \end{subfigure}\\
        \begin{subfigure}{\linewidth}
            \centering
            \includegraphics{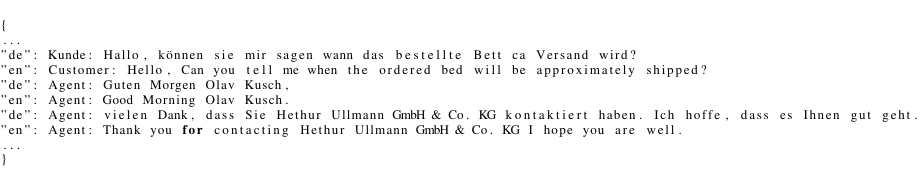}
         \caption{Sentence-level with artificial replaced PII training pairs}
            \label{fig:example-sen-maia}
     \end{subfigure}\\
             \begin{subfigure}{\linewidth}
                 \centering
            \includegraphics{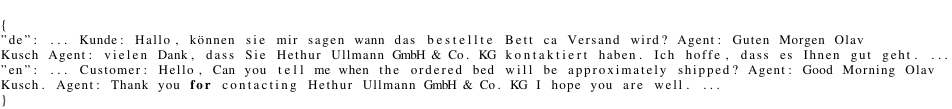}
         \caption{Document-level with artificial replaced PII training pair (utterances within a dialogue)}
     \end{subfigure}
     \caption{Difference between training examples for the document-level vs sentence-level MAIA dataset.}
    \label{fig:example-doc-vs-sen-maia}
\end{figure*}

Figure~\ref{fig:example-sen-maia} shows an example of the MAIA dataset with replaced PII data.

\paragraph{MAIA}
The preprocessing for the MAIA dataset is similar to the BSD dataset.
Moreover, since MAIA is a real-world dataset from Unbabel's client, it is anonymized before being released.
To make the dataset more realistic, we replace the anonymized PII data with artificial PII data.
We use \texttt{Faker}\footnote{\url{https://faker.readthedocs.io/en/}} to generate fake PII data and replace the pre-anonymized PII data in the MAIA dataset.
In each dialogue, we keep the replaced PII data consistent across all utterances (e.g., \#NAME\# is always replaced by one artificial name within a dialogue).
We also use localized fake data for each language 
(e.g.\ \#PRS\_ORG\# is replaced by a German company name).

\begin{table}[t]
\centering
 \resizebox{\columnwidth}{!}{%
\begin{tabular}{@{}ll@{}}
\toprule
Notation   & Description \\ \midrule
\dtrainsen & Sentence-level training data            \\
\dvalsen   & Sentence-level validation data            \\
\dtestsen  & Sentence-level test data            \\
\dtraindoc & Document-level training data            \\
\dvaldoc   & Document-level validation data            \\
\dtestdoc  & Document-level test data            \\
\senm      & Finetuned model on \dtrainsen            \\
\docm      & Finetuned model on \dtraindoc             \\
\zaugm     & Finetuned model on document-level WMT22            \\
\augm      & Finetuned model on \dtraindoc with~\zaugm~checkpoint           \\
\bottomrule
\end{tabular}}
\caption{Notation used in this work.}
\label{tab:notation}
\end{table}

\section{Hyperparameters}\label{sec:hyperparameters}
We first consider the optimal maximum sequence length.
For tokenization of the training data, we use the SentencePiece tokenizer~\citep{kudo-richardson-2018-sentencepiece} that comes with mLongT5.
The tokenizer is trained on mC4~\citep{colin-2020-t5} with a vocabulary size of 256,384.
As shown in Table~\ref{tab:my-table-stats-token}, all datasets have a long tail distribution of token length.

Regarding hyperparameter tuning, we divide this into two cases: (1) Normal training and (2) private training.
For both cases, we always set the maximum sequence length to the longest sequence in the training dataset.
We conducted an experiment with truncated sequences at 512, 256, and 128 tokens on \docm.
The results of 512 and 256 token sequence lengths are better than setting the model to the longest sequence in the training data, when using a high value of $\varepsilon$.
However, the results at small $\varepsilon$ values is indifferent to that reported in this work at the longest sequence of the training data.

\begin{table*}[!t]
\centering
\begin{tabular}{@{}llccccccccc@{}}
\toprule
     &          & \multicolumn{3}{c}{Train} & \multicolumn{3}{c}{Validation} & \multicolumn{3}{c}{Test} \\
     &          & $\mu$  & $\sigma^2$& $\max$ & $\mu$   & $\sigma^2$  & $\max$   & $\mu$ & $\sigma^2$& $\max$ \\ \midrule
BSD  & Japanese & 491    & 165     & 1007   & 484     & 155       & 843      & 495   & 150     & 1025   \\
     & English  & 499    & 174     & 1090   & 486     & 163       & 870      & 495   & 158     & 1060   \\
MAIA & German   & 589    & 278     & 1606   & 466     & 440       & 1101     & 515   & 488     & 1200   \\
     & English  & 555    & 262     & 1504   & 180     & 174       & 1034     & 242   & 230     & 1160   \\
 Europarl & German   & 402    & 397     & 10899   & 349     & 365       & 7077     & 351   & 368     & 6754   \\
     & English  & 371    & 368     & 10046   & 318     & 334       & 6923     & 320   & 336     & 5917   \\\bottomrule
\end{tabular}
\caption{Maximum token length, approximate mean and standard variation of each language in the document-level BSD and MAIA datasets. We use the SentencePiece tokenizer to tokenize the data.}
\label{tab:my-table-stats-token}
\end{table*}

\begin{table*}[!t]
    \resizebox{\textwidth}{!}{%
        \begin{tabular}{@{}lllllll@{}}
        \toprule
        Dataset & $\epsilon$  & Training Unit  & Max\text{.} seq\text{.} length & $lr$ & Epochs & Total Batch Size \\ \midrule
        MAIA    & $\infty$    & Sentence-Level & 128                 & $1e-3$          & 30     & 32               \\
                & $\infty$    & Document-Level & 1610                & $3e-3$          & 25     & 2                \\
                & $\{990,99,10,1\}$ & Sentence-Level & 128                 & $1e-2$          & 30     & 1024             \\
                & $\{990,99,10,1\}$ & Document-Level & 1610                & $1e-2$          & 100    & 256              \\
        BSD     & $\infty$    & Sentence-Level & 64                  & $1e-3$          & 15     & 32               \\
                & $\infty$    & Document-Level & 1100                & $3e-3$          & 30     & 2                \\
                & $\{400,40,10,1\}$ & Sentence-Level & 64                  & $1e-2$          & 15     & 1024             \\
                & $\{400,40,10,1\}$ & Document-Level & 1100                & $1e-2$          & 100    & 512              \\
        Europarl     & $\infty$    & Sentence-Level & 128                  & $1e-4$          & 16     & 128               \\
                & $\infty$    & Document-Level & 1500                & $1e-4$          & 30     & 128                \\
                & $\{3130,313,10,1\}$ & Sentence-Level & 128                  & $1e-3$          & 16     & 1,048,576             \\
                & $\{3130,313,10,1\}$ & Document-Level & 1500                & $1e-3$          & 100    & 131,072              \\
        WMT22-JA-EN & $\infty$    & Document-Level & 1200                & $1e-2$          & 2      & 4                \\
        WMT22-DE-EN & $\infty$    & Document-Level & 1500                & $1e-2$          & 2      & 4                \\ \bottomrule
        \end{tabular}
    }
\caption{Final results for hyperparameter search.}
\label{tab:training}
\end{table*}

\begin{table*}[!t]
    \begin{subtable}{.6\linewidth}
      \centering
        \begin{tabular}{lll}
        \toprule
        Dataset & \# Member & \# Non-member \\ \midrule
        BSD     & 4147      & 4147          \\
        MAIA    & 4597      & 4597          \\
        Europarl    & 363,538      & 363,538          \\
            \bottomrule
        \end{tabular}
        \caption{Number of examples used for sentence-level loss-based MIA.}
        \label{tab:my-table-mia}
    \end{subtable}%
    \begin{subtable}{.4\linewidth}
        \centering
        \begin{tabular}{lll}
            \toprule
            Dataset & \# PII \\ \midrule
            BSD     & 1286   \\
            MAIA    & 1156   \\
            Europarl    & 27,527   \\
            \bottomrule
        \end{tabular}
        \caption{Number of PII in $\mathcal{D}_{\mathrm{sampled}}^{\mathsf{sen}}$ of each dataset}
        \label{tab:my-table-pii}
    \end{subtable}\caption{Statistics of datasets used for MIA evaluation.}
    \label{tab:table}
\end{table*}

Table~\ref{tab:training} shows the final hyperparameters for each dataset.

\subsection{Hyperparameter tuning for normal training}
In normal training, we only use one seed for hyperparameter tuning, but three runs for each final selected hyperparameter to get the average performance. We conduct experiments on two H100 GPUs.

\begin{table}[t]
\centering
\begin{tabular}{@{}llll@{}}
\toprule
Lang. Pair & Sen.-level & Doc.-level \\ \midrule
JA-EN         & 33,875,119       & 851,525         \\
DE-EN         & 295,805,439      & 4,779,636        \\ \bottomrule
\end{tabular}\caption{Number of training examples in WMT22 dataset.}
\label{tab:tablewmt}
\end{table}

\subsection{Hyperparameter tuning for private training}
For the final results with DP-SGD, we run two trials for each hyperparameter configuration, with five different seeds for each final selected hyperparameter, for each $\varepsilon$ value. 
We use the same notation for \textit{epochs} when running DP-SGD training with Poisson sampling as \citet{abadi2016deep}, being $\frac{N}{L}$.
Next, we utilize very large batch sizes for both of these methods, setting $L$ to a large value and building up the resulting drawn batches with gradient accumulation.
All private training experiments are conducted using one H100 GPU, due to the limitation of duplicating examples in lots when sampling to multiple GPUs. 
Similar to normal training, we keep the same maximum sequence length for both datasets in private training.

\subsection{Augmented training for document-level models} \label{subsec:aug_training}

We then fine-tune the mLongT5 checkpoint on the concatenated documents for 2 epochs with a batch size of 16 on two H100 GPUs.
The hyperparameter search space is the same as for normal training, as described above.
For Japanese to English translation, we use the entire WMT dataset.
However, for German to English translation, we only use the first part of the dataset.
This is done to ensure that the dataset size is equal to that of the Japanese to English dataset (851,525), due to time constraints.\footnote{It takes 98 hours to finish one epoch on the entire document-level WMT22. We also split the dataset by half and trained the model on each part. The results are poor compared to using only a small part of the data, despite training of each part to simulate training the entire dataset in one epoch.}

\begin{figure*}[!tb]
        \centering
                \begin{subfigure}{\linewidth}
            \centering
                 \includegraphics{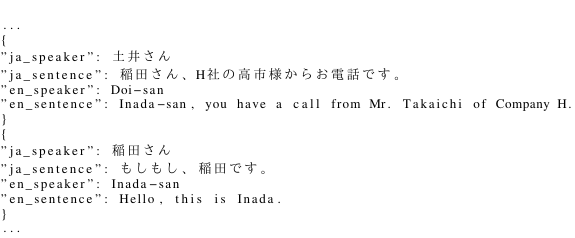}
         \caption{Sentence-level original training pairs}
     \end{subfigure}\\
        \begin{subfigure}{\linewidth}
            \centering
                 \includegraphics{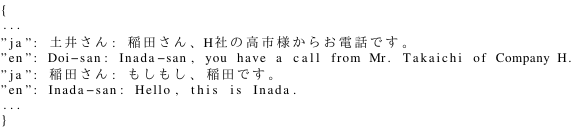}
         \caption{Sentence-level modified training pairs}
     \end{subfigure}\\
             \begin{subfigure}{\linewidth}
                 \centering
                 \includegraphics{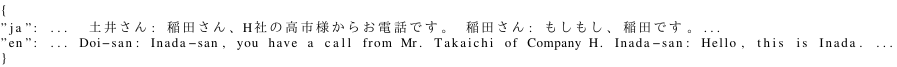}
         \caption{Document-level training pair (utterances within a dialogue)}
     \end{subfigure}
     \caption{Difference between training examples for the document-level vs sentence-level BSD dataset.}
    \label{fig:example-doc-vs-sen-bsd}
\end{figure*}

\begin{figure*}[!tb]
        \centering
                \begin{subfigure}{\linewidth}
            \centering
                 \includegraphics{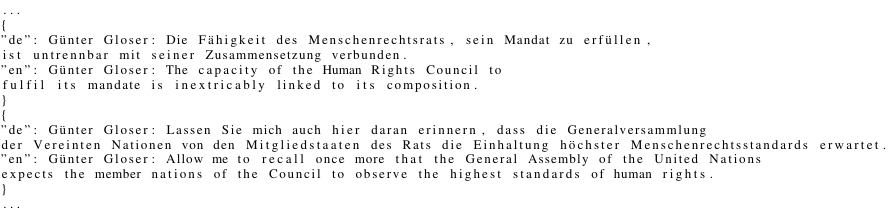}
         \caption{Sentence-level original training pairs}
     \end{subfigure}\\
             \begin{subfigure}{\linewidth}
                 \centering
                 \includegraphics{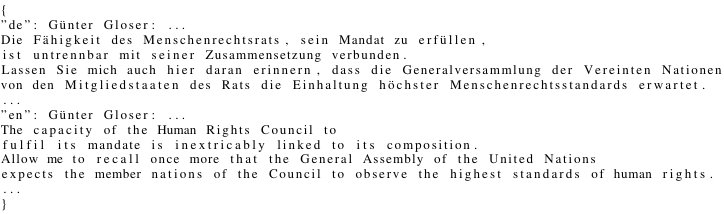}
         \caption{Document-level training pair (utterances within a speech)}
     \end{subfigure}
     \caption{Difference between training examples for the document-level vs sentence-level Europarl dataset.}
    \label{fig:example-doc-vs-sen-europarl}
\end{figure*}

\begin{figure*}[!t]
\centering
\begin{subfigure}{.33\textwidth}
  \centering
  \includegraphics[width=\linewidth]{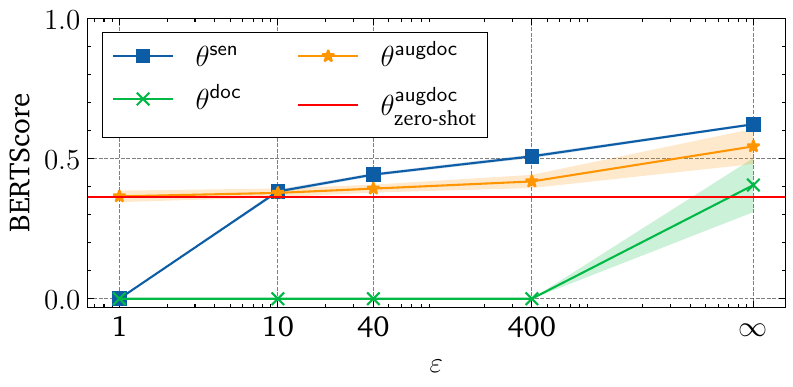}
  \caption{BSD}
  \label{fig:bertscore-sen-test-bsd}
\end{subfigure}
\begin{subfigure}{.315\textwidth}
  \centering
  \includegraphics[width=\linewidth]{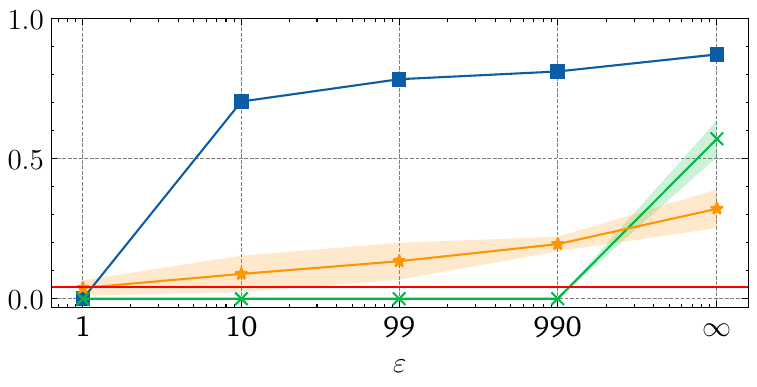}
  \caption{MAIA}
  \label{fig:bertscore-sen-test-maia}
\end{subfigure}%
\begin{subfigure}{.315\textwidth}
  \centering
  \includegraphics[width=\linewidth]{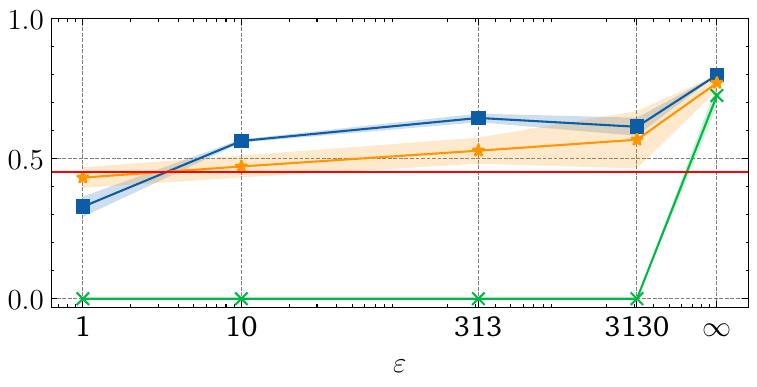}
  \caption{Europarl}
  \label{fig:bertscore-sen-test-Europarl}
\end{subfigure}%
\caption{BERTScore on \dtestsen~for all four model fine-tuning configurations. Lower $\varepsilon$ corresponds to better privacy.
}
\label{fig:bertscore-sen-test}
\end{figure*}

\begin{figure*}[!t]
\centering
\begin{subfigure}{.33\textwidth}
  \centering
  \includegraphics[width=\linewidth]{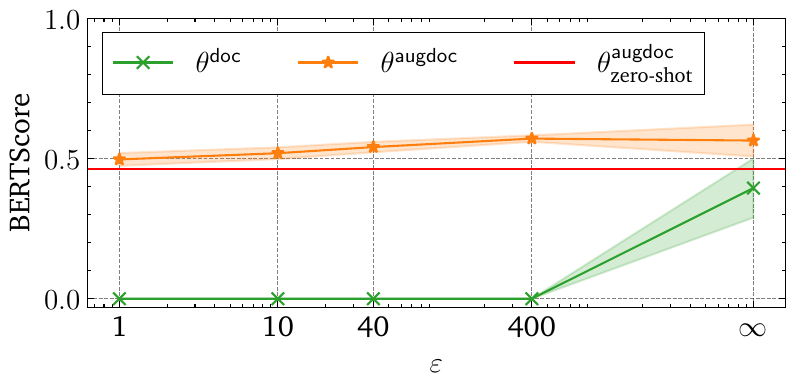}
  \caption{BSD}
  \label{fig:test122}
\end{subfigure}
\begin{subfigure}{.315\textwidth}
  \centering
  \includegraphics[width=\linewidth]{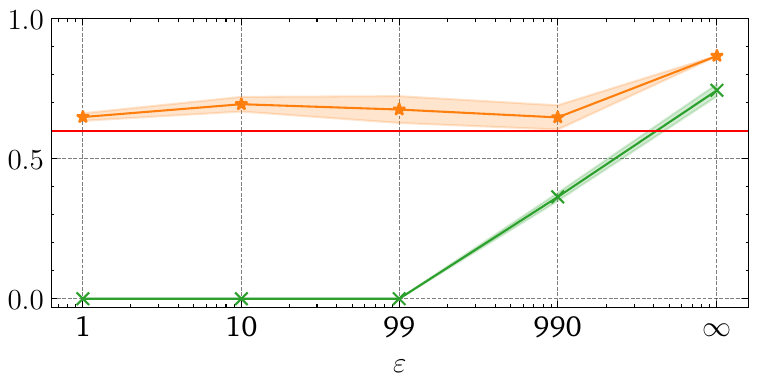}
  \caption{MAIA}
  \label{fig:test111}
\end{subfigure}%
\begin{subfigure}{.315\textwidth}
  \centering
  \includegraphics[width=\linewidth]{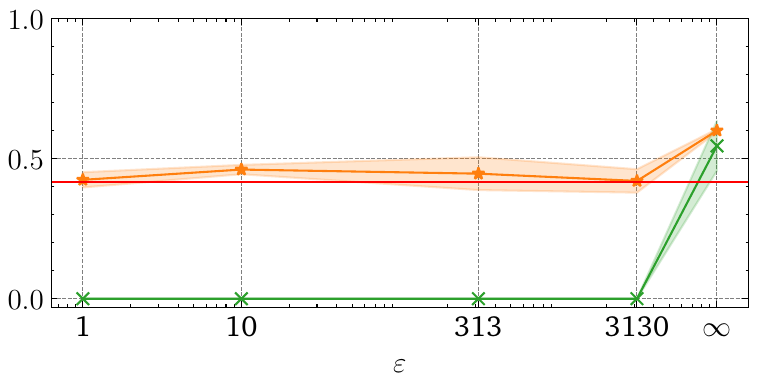}
  \caption{Europarl}
  \label{fig:test112}
\end{subfigure}%
\caption{BERTScore on \dtestdoc~for the three document-level model fine-tuning configurations. Lower $\varepsilon$ corresponds to better privacy.
}
\label{fig:bertscore-doc-test}
\end{figure*}

\section{BERTScore Modification} \label{bertscore-modification}
We also use BERTScore~\citep{zhang2019bertscore} for semantic similarity evaluation.
BERTScore uses RoBERTa embeddings \citep{liu2019roberta} to compute the similarity between the candidate translation and the reference translation.
However, its embeddings are limited to 512 tokens, which is not enough for our task.
Therefore, we modify BERTScore to use Longformer\footnote{\url{https://huggingface.co/allenai/longformer-base-4096}} embeddings~\citep{beltagy2020longformer} instead of RoBERTa embeddings.
Longformer's embeddings are able to encode long sentences up to 4,096 tokens.
Another modification that we make to BERTScore in this work is rescaling the score baseline to make it more readable.

The score is computed from the seventh layer output of Longformer's embeddings, since it has the best correlation with human judgment on the WMT16 Metrics Shared Task~\citep{zhang2019bertscore}.
BERTScore uses pre-normalized vectors for cosine similarity, resulting in computed scores the range $[-1, 1]$.
However, in practice, the observed BERTScore values are often limited to a narrow range~\citep{moosavi2021scigen, zhang2019bertscore}.
For instance, when using the default large RoBERTa\footnote{\url{https://huggingface.co/roberta-large}} model, BERTScore typically falls between 0.85 and 0.95.
This is due to the learned geometry of contextual embeddings which results in different scores from different embeddings.
Although this characteristic does not affect BERTScore's ability to rank text generation systems, it does make the resulting score less comprehensible to humans.
To address this issue, ~\citet{zhang2019bertscore} rescale BERTScore using its empirical lower bound $b$ as a baseline.
The computation of~$b$ is carried out using Common Crawl\footnote{\url{https://commoncrawl.org/}} monolingual datasets.
For each language and contextual embedding model, one million candidate-reference pairs are created by grouping two random sentences.
Due to the random pairing and corpus diversity, each pair has very low lexical and semantic overlap (BLEU computed on these pairs is around zero).
To compute the value of $b$, ~\citet{zhang2019bertscore} take the average of BERTScore computed on the sentence.
After that, using baseline $b$, we get a linearly rescaled BERTScore.

\begin{equation*}
    \hat{F}_{\mathsf{BERT}} = \frac{F_{\mathsf{BERT}} - b}{1 - b}
\end{equation*}

The result $\hat{F}_{\mathsf{BERT}}$ typically ranges between 0 and 1, with anything below this range clipped to 0.
As~\citet{zhang2019bertscore} note, this method does not affect the ranking ability or human correlation of BERTScore, as measured by Pearson's and Kendall's coefficients, but to enhance the readability of the score.
Since the Longformer rescaled BERTScore is not available, we compute it ourselves.

\section{MIA experiment details}\label{sec:stats-attack}
Table~\ref{tab:my-table-mia} shows the number of members and non-members for MIA evaluation.

\section{Privacy/utility trade-off evaluation with BERTScore}
Apart from using BLEU for evaluating the privacy/utility trade-off (Section~\ref{eval:performance}), we also use BERTScore for the evaluation of translation quality at the document level, with respect to semantic similarity.
\subsection{Privacy/utility trade-off on \dtestdoc}
Figure~\ref{fig:bertscore-doc-test} shows the BERTScore of~\docm~and~\augm~on~\dtestdoc.
It is evident that the BERTScore decreases as the value of~$\varepsilon$ is decreased for both datasets, similar to the BLEU score results.
Compared to BLEU scores, the main difference is that~\augm~is more stable across different values of~$\varepsilon$, in particular for the BSD dataset.
It is also worth noting that at $\varepsilon=400$ on BSD, the BERTScore of~\docm~is 0, while the BLEU score is still around 2.
On MAIA, the BERTScore of~\docm~and~\augm~shows less variation at $\varepsilon=\infty$ compared to the BLEU score. 
Regarding Europarl, the BERTScore of~\docm~is 0 when training with DP-SGD, while~\augm~performance trend is similar to BSD and MAIA results.

\subsection{Privacy/utility trade-off on \dtestsen}

Figure~\ref{fig:bertscore-sen-test} shows the BERTScore of~\senm, \docm~and~\augm~on~\dtestsen.

\paragraph{MAIA} On MAIA, the BERTScore of~\senm~slowly decreases as the value of~$\varepsilon$ is decreased, for about 0.1 points per decrease~in~$\varepsilon$ from $\infty$ to $10$.
It is interesting that the BERTScore of~\augm~varies a lot across different values of~$\varepsilon$, e.g., $0.088\pm0.079$ at $\varepsilon=10$ and $0.134\pm0.081$ at $\varepsilon=99$.

\paragraph{BSD} On BSD, the BERTScores of~\senm,~\docm~and~\augm~are very close to one other, especially at $\varepsilon=\infty$.
From $\varepsilon=400$, we no longer observe any meaningful translation quality~from~\docm.
The BERTScores of~\senm~and~\augm~start to converge and become equal at $\varepsilon=10$, which demonstrates that the semantic content~of~\senm~and~\augm~equals at the sentence level on the BSD dataset.
As expected, the BERTScore of~\senm~at $\varepsilon=1$ is equal to zero due to the high amount of added noise during the training process with DP-SGD, while the BERTScore of~\augm~is equal to that of~\zaugm~at approximately 0.38.
\label{sec:appx-detailed-results}

\paragraph{Europarl}
On Europarl, the BERTScores of~\senm,~\docm~and~\augm~are even closer at $\varepsilon=\infty$ than on BSD. While at $\varepsilon=3130$~\docm~fails to generate a sentence,~\senm~and~\augm~have similar drops in their performance. Thanks to the large amount of training data,~\senm~keeps the results at 0.4 BERTScore with $\varepsilon=1$. Unlike the BLEU score results,~\senm~impressively maintains the semantic similarity of the generated sentence with 30\% reduction.

\begin{table*}[!t]
  \resizebox{\textwidth}{!}{%
\begin{tabular}{@{}lll@{}}
\toprule
Model                        & $\epsilon$   & System Output                                                                                                                                                                                                                                                                                                                                                                                                                                                                                                                                                                   \\ \midrule
\senm              & $\infty$     & Customer: I just bought a book with my Geisler Conradi GmbH, it seems to be on it.                                                                                                                                                                                                                                                                                                                                                                                                                                                                                              \\
                        & $990$       & Customer: I bought a book directly with my Geisler Conradi GmbH, it seems to be on it.                                                                                                                                                                                                                                                                                                                                                                                                                                                                                          \\
                        & $99$        & Customer: I bought a book directly with my Geisler Conradi GmbH, it seems to be on it.                                                                                                                                                                                                                                                                                                                                                                                                                                                                                          \\
                        & $10$        & Customer: I bought the Geisler Conradi GmbH, a book it seems to be on.                                                                                                                                                                                                                                                                                                                                                                                                                                                                                                          \\\midrule
\docm              & $\infty$     & Customer: I bought a book directly with my Geisler Conradi GmbH, it seems to be on it.                                                                                                                                                                                                                                                                                                                                                                                                                                                                                          \\\cmidrule(l){2-3}
                        & $990$       & \begin{tabular}[c]{@{}l@{}}Customer: I ordered directly with Geisler Conradi GmbH, a book. It seems to be on..\\ \textbackslash{}n\\ Customer: I ordered directly with my Geisler Conradi GmbH, a book. It seems to be on..\\ \textbackslash{}n\\ Customer: I ordered directly with my Geisler Conradi GmbH, a book. It seems to be on..\\ \textbackslash{}n\\ Customer: I ordered directly with my Geisler Conradi GmbH, a book. It seems to be on..\\ \textbackslash{}n\\ Customer: I ordered directly with my Geisler Conradi GmbH, a book. It seems to be on..\end{tabular} \\ \midrule
\augm           & $\infty$     & Customer: I bought a book directly with my Geisler Conradi GmbH and it seems to be on it.                                                                                                                                                                                                                                                                                                                                                                                                                                                                                       \\\cmidrule(l){2-3}
                        & $990$       & \begin{tabular}[c]{@{}l@{}}Customer: I bought directly with my Geisler Conradi GmbH, a book it seems to be on it.\textvisiblespace \\ Customer: I bought directly with my Geisler Conradi GmbH, a book it seems to be on it.\textvisiblespace \\ Customer: I bought directly with my Geisler Conradi GmbH, a book it seems to be on it.\textvisiblespace \\ Customer: I bought directly with my Geisler Conradi GmbH, a book it seems to be on it.\textvisiblespace \\ Customer: I bought directly with my Geisler Conradi GmbH, a book it seems to be on it.\end{tabular}                                                                          \\\cmidrule(l){2-3}
                        & $99$        & \begin{tabular}[c]{@{}l@{}}Customer: Have directly with my Geisler Conradi GmbH, bought a book it seems to be on it too.\textvisiblespace \\ Customer: Have directly with my Geisler Conradi GmbH, bought a book it seems to be on it too.\textvisiblespace \\ Customer: Have directly with my Geisler Conradi GmbH, bought a book it seems to be on it too.\textvisiblespace \\ Customer: Have directly with my Geisler Conradi GmbH, bought a book it seems to be on it too.\textvisiblespace \\ Customer: Have directly with my Geisler Conradi GmbH, bought a book it seems to be on it too.\end{tabular}                                       \\\cmidrule(l){2-3}
                        & $10$        & \begin{tabular}[c]{@{}l@{}}Customer: Have directly with my Geisler Conradi GmbH, a book bought it seems to be on it too. \\ Customer: Have directly with my Geisler Conradi GmbH, a book bought it seems to be on it too.\textvisiblespace \\ Customer: Have directly with my Geisler Conradi GmbH, a book bought it seems to be on it too.\textvisiblespace \\ Customer: Have directly with my Geisler Conradi GmbH, a book bought it seems to be on it too. \\ Customer: Have directly with my Geisler Conradi GmbH, a book bought it seems to be on it too.\end{tabular}                                       \\\cmidrule(l){2-3}
                        & $1$         & \begin{tabular}[c]{@{}l@{}}Customer: Have directly with my Geisler Conradi GmbH, a book bought it seems to be on it.\textvisiblespace \\ Customer: Have directly with my Geisler Conradi GmbH, a book bought it seems to be on it.\textvisiblespace \\ Customer: Have directly with my Geisler Conradi GmbH, a book bought it seems to be on it.\textvisiblespace \\ Customer: Have directly with my Geisler Conradi GmbH, a book bought it seems to be on it.\textvisiblespace \\ Customer: Have directly with my Geisler Conradi GmbH, a book bought it seems to be on it.\end{tabular}                                                           \\ \midrule
\zaugm & -         & \begin{tabular}[c]{@{}l@{}}Customer: Have directly with my Geisler Conradi GmbH, a book bought it seems to be on it too.\textvisiblespace \\ Customer: Have directly with my Geisler Conradi GmbH, a book bought it seems to be on it too.\textvisiblespace \\ Customer: Have directly with my Geisler Conradi GmbH, a book bought it seems to be on it too.\textvisiblespace \\ Customer: Have directly with my Geisler Conradi GmbH, a book bought it seems to be on it too.\textvisiblespace \\ Customer: Have directly with my Geisler Conradi GmbH, a book bought it seems to be on it too.\end{tabular}                                       \\ \midrule
 Original               & Reference & Customer: I bought a book directly with my Geisler Conradi GmbH, it seems to be on it too.                                                                                                                                                                                                                                                                                                                                                                                                                                                                                      \\
                        & Source    & Kunde: Habe direkt mit meinem Geisler Conradi GmbH, ein buch gekauft es scheint auch drauf zu sein.                                                                                                                                                                                                                                                                                                                                                                                                                                                                             \\ \bottomrule
\end{tabular}}
\caption{Translation sample from~\dtestsen~of MAIA dataset}
\label{tab:dtestsen-maia-sample}
\end{table*}

\begin{table*}[!t]
    \includegraphics[width=\linewidth]{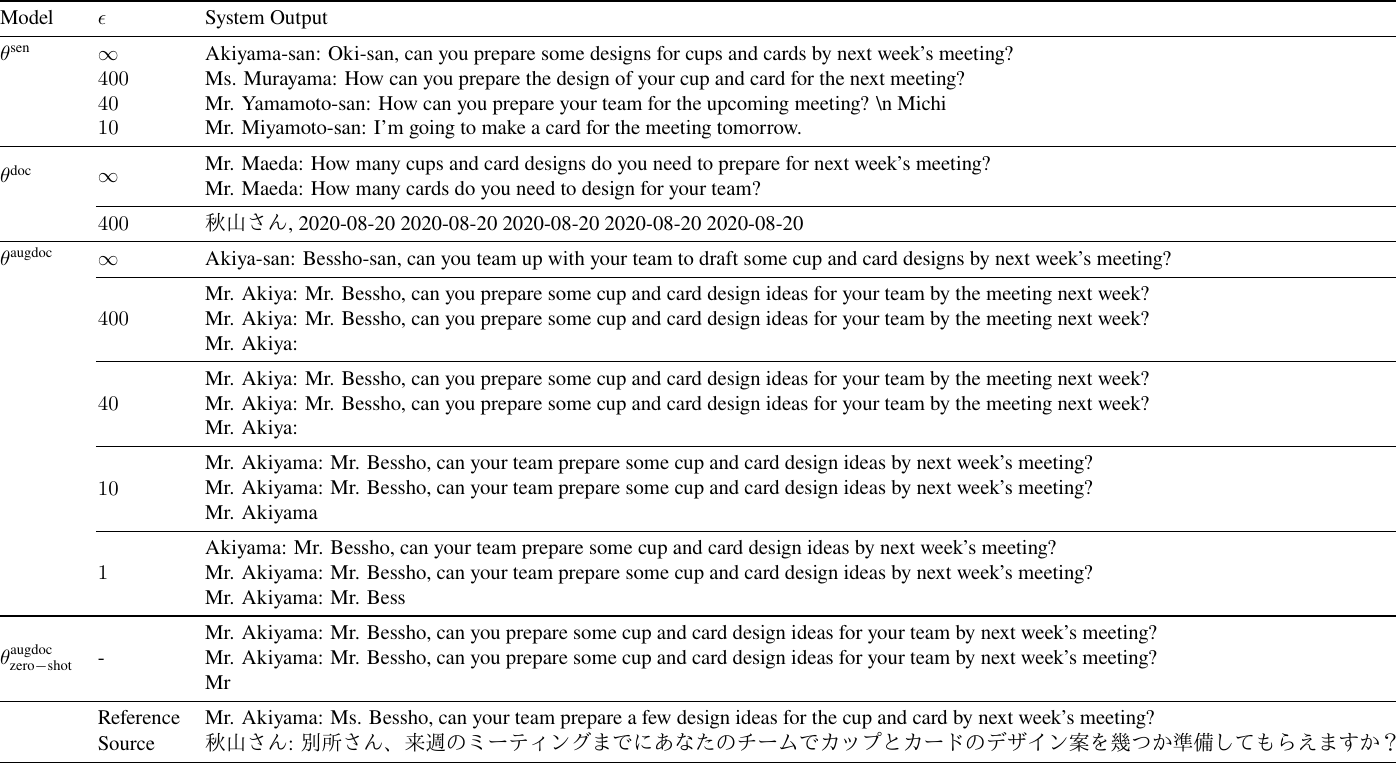}
\caption{Translation sample from~\dtestsen~of BSD dataset}
\label{tab:dtestsen-BSD-sample}
\end{table*}

\begin{table*}[!t]
    \includegraphics[width=\linewidth]{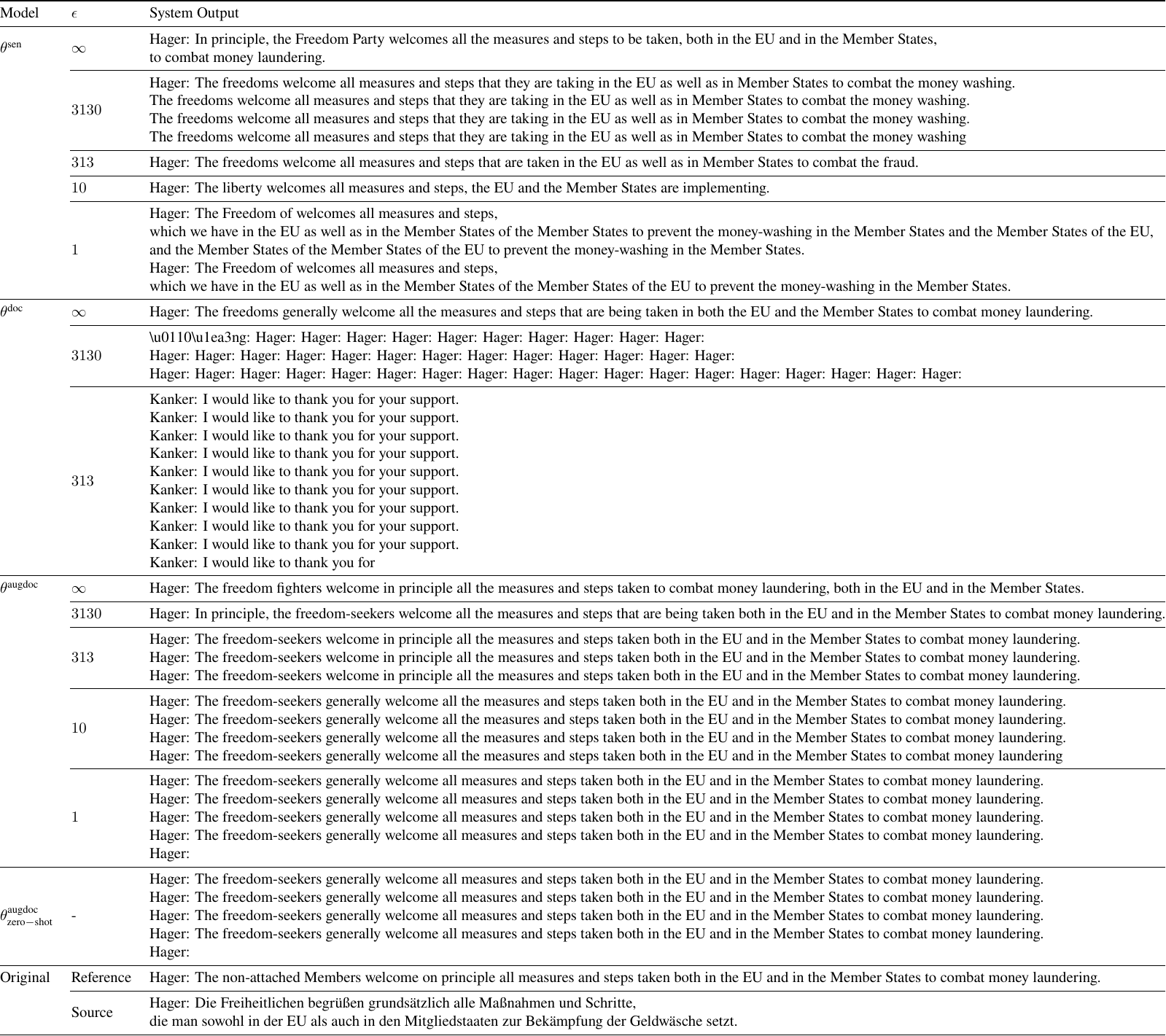}
\caption{Translation sample from~\dtestsen~of Europarl dataset}
\label{tab:dtestsen-Europarl-sample}
\end{table*}

\section{Discussion on translation quality}\label{sec:more_discussion}
Finally, Table~\ref{tab:dtestsen-maia-sample} and Table~\ref{tab:dtestsen-BSD-sample} show that the document-level training model mainly duplicates the sentence until it reaches the maximum sequence length.
Overall,~\augm~shows better performance for private training than~\docm~and is close to~\senm~performance with post-processing.

\section{DP guarantees in our experiments}\label{sec:our-privacy-guarantees}

To provide all the information needed to understand our privacy guarantees, we follow the guidelines
outlined in~\citet{ponomareva2023dp}.

\begin{enumerate}
    \item \textbf{DP setting.} We provide a central DP guarantee where the service provider is trusted to correctly implement the mechanism.
    \item \textbf{Instantiating the DP Definition}
    \begin{enumerate}
        \item \emph{Data accesses covered}: Our DP guarantees apply only to a single training run.
        We don't account for hyperparameter tuning in our guarantees.
         Public multilingual C4 data~\citep{colin-2020-t5, xue-etal-2021-mt5} is used for pre-training mLongT5.
        \item \emph{Final mechanism output}: Only the model predictions, such as the translated sentences generated by the models trained with DP, are released.
        The mechanism's output is technically the full sequence of privatized gradients, and the guarantee also applies at this level.
        Hence, all checkpoints are protected and can be released publicly.
        \item \emph{Unit of privacy.} Since we are working in the NLP context, we consider sentences and documents as the unit of privacy.
        The sentence-level unit is an utterance in a conversation, typically a single sentence with a maximum length of 64 to 128 tokens, depending on the dataset.
        The document-level unit is the whole conversation dialogue, which can be composed of multiple sentences.
        Thus, the maximum length of the document is not limited, and in our experiments, the maximum length of the document is up to 1,700 tokens.
        Token counting is done after tokenization using SentencePiece~\citep{kudo-richardson-2018-sentencepiece}.
        We demonstrate in our experiments that sentence-level privacy is weaker than document-level privacy.
        However, group privacy can be used to achieve document-level privacy from sentence-level privacy.
        \item \emph{Adjacency definition for ``neighboring'' datasets}: We use the add-or-remove adjacency definition.
    \end{enumerate}
    \item \begin{enumerate}
        \item \textit{Type of accounting used}: RDP-based accounting.
        \item \textit{Accounting assumptions}: We correctly use Poisson sampling.
        \item \textit{The formal DP statement}: We use various levels of $\varepsilon$~values:~$1, 10, 40, 99, 400, 990$.
        Our $\delta$ is set to $10^{-8}$.
        \item \textit{Transparency and verifiability}: We are going to open source our code based on the open-source DP-NMT framework~\citep{igamberdiev-etal-2024-dp}.
    \end{enumerate}
\end{enumerate}

\end{document}